\documentclass{article}[letter]
\PassOptionsToPackage{sort&compress}{natbib} 
\usepackage[final]{neurips_2023}
\usepackage[utf8]{inputenc} 
\usepackage[T1]{fontenc}    
\usepackage{hyperref}       
\usepackage{url}            
\usepackage{booktabs}       
\usepackage{amsfonts}       
\usepackage{amsmath}        
\usepackage{nicefrac}       
\usepackage{microtype}      
\usepackage{xcolor}         

\setcitestyle{numbers}

\usepackage{graphicx}
\usepackage[most]{tcolorbox}
\usepackage{quoting, xparse}    
\usepackage[bb=boondox]{mathalfa}
\usepackage{comment}
\usepackage{fancyhdr}
\newcommand{\examplebox}[2]{
  \vspace{.25em}
  \begin{tcolorbox}[
    width=\columnwidth,
    colback=gray!30!white,
    colframe=black]
    \textbf{#1} \\
    #2
  \end{tcolorbox}
}
\renewcommand{\examplebox}[2]{
  \begin{tcolorbox}[
    breakable,
    adjusted title=#1,
    fonttitle=\bfseries,
    attach title to upper={\\\\},
    coltitle=black,
    before upper=\setlength{\parskip}{0.5em}{\tcbtitle\par},
    width=\columnwidth,
    colback=gray!30!white,
    colframe=black]
    \renewcommand{\\}{} 
    #2
  \end{tcolorbox}
}

\NewDocumentCommand{\bywhom}{m}{
  {\nobreak\hfill\penalty50\hskip1em\null\nobreak
   \hfill\mbox{\normalfont(#1)}%
   \parfillskip=0pt \finalhyphendemerits=0 \par}%
}

\NewDocumentEnvironment{pquotation}{m}
  {\begin{quoting}[
     indentfirst=true,
     leftmargin=\parindent,
     rightmargin=\parindent]\itshape}
  {\bywhom{#1}\end{quoting}}

\title{A Guide to Failure in Machine Learning: Reliability and Robustness from Foundations to Practice}
\author{%
  Eric Heim, Oren Wright, David Shriver \\
  Software Engineering Institute\\
  Carnegie Mellon University\\
  \texttt{\{etheim, owright, dlshriver\}@sei.cmu.edu} \\
}

\begin{document}
\maketitle
\thispagestyle{fancy}

\begin{abstract}
  One of the main barriers to adoption of Machine Learning (ML) is that ML models can fail unexpectedly.
  In this work, we aim to provide practitioners a guide to better understand why ML models fail and equip them with techniques they can use to reason about failure.
  Specifically, we discuss failure as either being caused by lack of \emph{reliability} or lack of \emph{robustness}.
  Differentiating the causes of failure in this way allows us to formally define why models fail from first principles and tie these definitions to engineering concepts and real-world deployment settings.
  Throughout the document we provide 1) a summary of important theoretic concepts in reliability and robustness, 2) a sampling current techniques that practitioners can utilize to reason about ML model reliability and robustness, and 3) examples that show how these concepts and techniques can apply to real-world settings.
\end{abstract}

\begin{pquotation}{Henry Petroski}
  \textit{Failure is central to engineering. Every single calculation that an engineer makes is a failure calculation. Successful engineering is all about understanding how things break or fail.}
\end{pquotation}

\section{Introduction}
\label{sec:intro}

Despite the increasing number of problems where Machine Learning (ML) surpasses the state-of-the-art, there has been a degree of reluctance to accept ML models as solutions in high-risk settings.
This general lack of trust in ML can be attributed in part to end-users and developers alike not understanding how, when, and why ML models \emph{fail}.
A number of well-known AI system failures such as an autonomous vehicle failing to identify a pedestrian~\cite{aiid:4}, and a smart smoke detector inadvertently silencing legitimate alarms~\cite{aiid:46} can be attributed to an ML model failing in unexpected and unhandled ways.
These and other cases highlight that need for clear understanding of when and how ML models can fail, so that the effects of ML failure can be mitigated and harm can be avoided.

In this guide, we seek to equip practitioners with both a basic understanding of the fundamentals of ML model failure, and methods to reason about it.
We do this by framing ML model failure in terms of \emph{reliability} and \emph{robustness}, two concepts commonly used in engineering disciplines but inconsistently used in ML literature.
Viewing failure under this lens allows us to formally define kinds of failures and map them to engineering concepts that are grounded in practice.
We discuss a sampling of current techniques to reason about failure and use formal definitions to scope when and how to use them.
As such, this guide is meant to provide a link from current formal understanding of failure in ML to practical guidance for developers\footnote{Throughout this guide we will often use ``developer'' as a blanket term for someone who has some role in building a system that includes machine learning.  The breadth of topics covered here range from design through fielding of systems, and in reality will involve many people with specific responsibilities and titles that cannot accurately be captured by ``developer''.  We choose not to focus on such organizational considerations in this document, but acknowledge their importance.} on how to build reliable and robust models.

With this in mind, we take an approach that emphasizes intuition over precision, and breadth over depth.
In Section \ref{sec:preliminaries}, we begin by going from an intuitive description of important concepts in machine learning to formal ML theory that can provide insight into how and why ML models fail.
In Section \ref{sec:reliability}, we provide a definition of ML model reliability and tie it to formalism introduced in the previous section.
From these, we then discuss some practical techniques to measure and possibly mitigate failures in reliability.
In Section \ref{sec:robustness}, we similarly define ML model robustness and contrast it to reliability.
As with reliability, we discuss foundations of ML model robustness and then review some basic techniques for reasoning about it.
Finally, in Section \ref{sec:conclusions} we conclude and highlight topics relevant to failure of ML models that were not covered in this guide.
Throughout the document we provide important additional references that can help fill technical points made in the paper, and a running example to showcase how the concepts we discuss apply to a real-world scenario
\footnote{Many key details about the design and implementation of the approaches highlighted in the running example are omitted.  As such, the examples should not be used as exact procedures to be followed, but illustrations of how to begin practically thinking about the concepts introduced in this guide.}.

\section{Preliminaries}
\label{sec:preliminaries}
To begin our discussion of ML model failure and how to prevent it, in this section we review some of the formal basis that describes the foundational assumptions in which ML models are built as a starting point for discussing how and why ML models can fail.
First, we provide basic terminology and formal definitions of the ML setting that is the focus of this work: Supervised learning.
Then, we briefly discuss key theoretic concepts that we will use to ground discussion going forward.

\subsection{Basic Definitions in Supervised Learning}\label{sec:formalism_supervised_learning}
Most common real-world ML problems can be broadly categorized as \emph{supervised learning} problems.
Supervised learning is the task of creating a \emph{model} that maps inputs to outputs from examples through a process called \emph{training}.
Intuitively, one can think of the model as a way to take data and make a \emph{prediction} about that data.
For instance, tasks such as determining whether a patient has a disease from their health record can be posed as a supervised learning problem.
One could collect data in the form electronic health records for patients, and then ask clinicians to provide feedback indicating which of the patients do and do not have the disease.
Patterns in this data can be found and exploited by a model to predict whether patients have the disease when it is given their electronic health record.

Formally, in supervised learning a model is defined as a function $f: \mathcal{X} \rightarrow \mathcal{Y}$ that maps inputs $x \in \mathcal{X}$ (often called \emph{instances}) to outputs $y \in \mathcal{Y}$ (often called \emph{labels}).
The distinction between broad classes of supervised techniques typically hinges on the choice for $\mathcal{Y}$.
For instance, if $\mathcal{Y} = \mathbb{R}$ (i.e. the model maps an instance to a real-valued number), then the problem being solved is called \emph{regression}.
If $\mathcal{Y} = \left\{c_1, c_2, ..., c_n\right\}$, where $c_i$ is a discrete category label, then the problem being solved is called $n$-\emph{way classification}\footnote{The disease prediction example can be defined as a 2-way (binary) classification problem where $\mathcal{Y} = \{disease\mathrm{,\ } no\ disease\}$.}.
We assume that $f$ will be selected from a \emph{hypothesis space} of possible functions $\mathcal{F}$.
Choosing the optimal model from $\mathcal{F}$ that correctly maps instances to their corresponding labels is the central challenge of supervised learning.

To choose the optimal model, training algorithms require a means to quantify how good a model is at mapping instances to labels.
For this, learning techniques employ a loss function (sometimes called a cost function or error function), $\ell: \mathcal{Y} \times \mathcal{Y} \rightarrow \mathbb{R}$.
This loss function is meant to produce a number that measures how close two labels are to each other for the purpose of comparing a model's output to true labels.
For example, the popular squared error loss used for regression problems is given by:
\begin{equation}\label{eq:squared_error}
    \ell(y, f(x)) = \left( f(x) - y \right)^2.
\end{equation}
Intuitively, for a given regression model $f$, \eqref{eq:squared_error} is low when the output of a model is close to a correct label $y$, as measured by the squared difference between a model's output and a correct label.

Equipped with a loss function, the most common theoretical assumption used in supervised learning is that an optimal model is one that minimizes \emph{expected error} (sometimes called \emph{risk}):
\begin{equation}\label{eq:risk}
    \mathbb{E}\left[ \ell(y, f(x)) \right] = \int_{\mathcal{X},\mathcal{Y}} \ell(y, f(x)) P(x,y) dxdy
\end{equation}
Equation \eqref{eq:risk} is the average loss of a model $f$ over all possible instance-label pairs, weighed by the probability of each pair occurring.
Choosing the model $f$ from $\mathcal{F}$ that minimizes \eqref{eq:risk} will give you the optimal
model in the hypothesis space, when considering all possible instances and labels and taking into account the probability of those instances and labels appearing as pairs for a given application.

For almost any real-world application, exactly computing \eqref{eq:risk} is impractical, since it is often impossible to collect every instance and label a model could encounter.
Instead, models are chosen using an approximation of \eqref{eq:risk} defined over a finite set of \emph{training data}:
\begin{equation*}\label{eq:training_data}
    \mathcal{D} = \left\{ (x_1, y_1), (x_2, y_2), \ldots, (x_n, y_n) \right\}.
\end{equation*}
This approximation of expected error using training data is called \emph{empirical error}:
\begin{equation}\label{eq:empirical_risk}
    \mathbb{E}\left[ \ell(y, f(x)) \right] \approx \frac{1}{n} \sum_i^n \ell(y_i, f(x_i))
\end{equation}
and finding a model (a process called \emph{training}) is most commonly done using specialized \emph{training algorithms} by choosing the model in the hypothesis space of models that minimizes empirical error:
\begin{equation}\label{eq:min_empirical_risk}
    \min_{f\in\mathcal{F}} \frac{1}{n} \sum_i^n \ell(y_i, f(x_i)).
\end{equation}

In summary, ``learning'' amounts to applying a training algorithm that utilizes training data to select a model from a hypothesis class by minimizing \eqref{eq:min_empirical_risk}.
For example, in regression, linear models are a popular choice where models are defined as: $f\left(x\right) = wx + b$.
The hypothesis space of such linear models are all possible settings of $w$ and $b$.
A training algorithm would choose settings for $w$ and $b$ such that empirical error is minimized over a given training set.

\examplebox{Running Example: Autonomous Vehicle with Visual Obstacle Avoidance}{As a running example, consider an autonomous vehicle meant to navigate city roads.
In order for such a system to be successful, it must detect obstacles in its path and avoid them.
The problem of detecting obstacles can be (and often is) posed a supervised learning problem.
While driving, the vehicle can use a camera to observe the area around the vehicle at regular intervals.
The images from the camera can be input into an ML model, and the model can predict the location of objects in the potential path of the car.
The vehicle can then use these predictions to avoid obstacles.

In the standard supervised learning setting, a team of developers would need to take a camera and collect instances that can be used for training the obstacle detection model.
They would also label the instances with the locations of the obstacles that the car would need to detect.
A training algorithm could then take the collected labeled training data and produce a model that would map images to obstacle locations.
Throughout the remainder of this document we will focus on a fictional team (called the obstacle avoidance team) tasked with developing such a supervised model to illustrate some important concepts more concretely
\footnote{Building an obstacle detection model that is reliable and robust enough to field on an autonomous system requires addressing a staggering number of important practical considerations.  We note this as our running example often over-simplifies these considerations to more succinctly emphasize the points made in this guide, and more careful consideration of the realities associated with autonomous driving is vital in practice.}}

\subsection{Generalization}
\label{sec:generalization}
The intent behind solving \eqref{eq:min_empirical_risk} is that the resulting model will \emph{generalize}, that is the model learned by minimizing empirical risk will output the correct labels for instances not in the training set\footnote{One important practical note: While the basic fundamental goal of machine learning is to learn a function $f$ that reduces expected error as measured with some ideal $\ell$, it is often the case that $\ell$ is not directly used during training for computational reasons (e.g. $\ell$ might not be ammenable to efficient training algorithms.).  In these cases, surrogate losses are used in \eqref{eq:min_empirical_risk}.}.
%
A significant body of theoretic machine learning research is dedicated to reasoning about the ability of ML models to generalize, as generalization from training data to data seen in deployment is perhaps the most important goal of ML.
Giving a full treatment of generalization is out of scope for this work
\footnote{Even providing a comprehensive list of references for the topic of generalization is a daunting task.  However, you can find formal introductions to the basics of generalization in most introductory machine learning textbooks.  See sections 5.2 and 5.4 of \cite{goodfellow2015} for one such discussion.}.
Instead of summarizing all important theoretical results in generalization, we wish to tie common important factors in the theory of generalization to things practitioners can do to ensure reliability and robustness.
In pursuit of this goal, we can broadly summarize work on generalization as focus on one or more of the following considerations:
\begin{enumerate}
    \item \textbf{Training Data} - What and how much training data is required to learn an accurate model.
    \item \textbf{Hypothesis Spaces} - What sets of models are amenable to learning models that generalize well under what conditions.
    \item \textbf{Training Algorithm} - What algorithms can be used to learn models that generalize.
\end{enumerate}
When building supervised models, developers can control each of these.
They determine what training data they collect, which class of model they use, and what training algorithm they employ
\footnote{Generally, learning theory often also considers some way of measuring the ``difficulty'' of a learning problem, but this is largely application dependent and outside the control of the practitioner.}.

As a classic example of how these can influence generalization, consider the concept of VC-dimension from Vapnik–Chervonenkis Theory~\cite{vapnik2013nature}.
Let $Err_{exp}$ be the expected error as defined in \eqref{eq:risk}, $Err_{emp}$ be the empirical risk as defined in \eqref{eq:empirical_risk}, and $VC\left(\mathcal{F}\right)$ be the VC-dimension 
\footnote{VC-Dimension can intuitively be understood as a way of measuring the representational power of a hypothesis space.  Generally speaking, more complex functions within a hypothesis space imply a larger VC-dimension.}
of the hypothesis space.
The VC-Dimension leads to reasoning about the relationship between empirical and expected error:
\begin{equation}
    Err_{exp} \leq Err_{emp} + VC\left(\mathcal{F}\right)
    \label{eq:VC-bound}
\end{equation}
Here, we can see the expected error of a model is no more than the sum of its empirical error and a particular way of measuring the hypothesis space of a learned model.
Not only does this provide a powerful way to reason about the way models can generalize, but it also highlights how choices made by developers affect generalization to data outside of the training set.
The hypothesis space of models is chosen by developers building a model, which is measured by the VC-dimension, and the empirical error is a function of all three choices listed above.

In practice, those building ML models will often not use theoretical results, such as \eqref{eq:VC-bound}, directly.
Despite this, many theoretical results are used as motivation for common practices in training and evaluating ML models.
For practitioners, we emphasize two important points regarding much of the theory explaining ML model generalization.
First, many of the most important theoretical results reason about error \emph{in expectation} or on average.
Even if one can prove that a model has low but non-zero expected error, it does not give insight as to \emph{what cases} the model will make errors.
In safety-critical applications, it is important to understand the specific conditions in which a model will fail, so safeguards can be put in place to mitigate harm.

Second, most theoretic results that explain generalization assume that the data used to train a model is drawn from the same distribution as the data the model observes during deployment.
Stated more concisely, theoretic generalization results assume that training and deployment data are \emph{identically distributed} (ID).
The ID assumption formalizes a relationship between training data and the data a model will observe when deployed.
It provides a connection that makes what a model learns during training applicable to a deployment task.
In the subsequent sections we will illustrate why this assumption is of vital importance to understanding and how to mitigate ML model failure.






%

\examplebox{Running Example: Autonomous Vehicle with Visual Obstacle Avoidance}{To identify obstacles, the obstacle avoidance team chooses to build a particular kind of supervised ML model called an \emph{object detector}, which is a model trained to take in an image as instance and predict so-called ``bounding boxes'' indicating the location of objects within the image, as well as labels indicating what each detected object is.
In this way, object detection can be seen as a combination of regression (pixel values indicating \emph{where} objects are; often called \emph{localization} of objects), and classification (class labels indicating \emph{what} objects are).

Specifically, they choose to learn from a popular class of models called You Only Look Once (YOLO)\cite{redmon2016you} object detectors.
The hypothesis space of models is defined by a \emph{neural network} architecture -- A nonlinear function characterized by a large number of free parameters to be set by a training algorithm -- designed for object detection.
YOLO models are trained using a loss function designed to balance errors in bounding box position, bounding box size, classification, and terms meant to discourage the model from detecting objects where there are none.
To train a YOLO model, the team utilizes a form \emph{gradient descent}, which are algorithms commonly employed to find a setting of neural network parameters that minimizes loss over training data.
If assumptions made by the hypothesis space, data, and algorithm hold, then the model should be able to generalize so that it detects obstacles to avoid during deployment.\\

While there has been considerable work on understanding theoretical properties of neural networks
\footnote{Again, a survey on theoretical results for neural networks is worthy of its own treatment outside of this work, but as examples \cite{bartlett2019nearly}, \cite{golowich2018size}, and \cite{du2019gradient} focus on VC-dimension, sample complexity, and gradient descent convergence properties, respectively.  \cite{suh2024survey} provides a recent survey on theoretical results of neural networks from a statistical perspective.}
that has informed the techniques used to build YOLO models, they cannot directly explain when a YOLO model will and will not fail in many practical scenarios.
For instance, if a practitioner trains an YOLO model for obstacle avoidance, there is generally no way to guarantee that when given an image, the model will always detect a pedestrian.
Considering the safety implications of failing to detect pedestrians in the path of an autonomous vehicle, it is important for the team to find other means to reason about how a model will generalize during deployment.

}

\subsection{A Definition of ML Model Failure}
Expected error is useful as a means to reason about the generalization performance of supervised models for broad classes of problem, but it alone does not define ML model failure in practice.
For that, we require additional information about the application in which the model is used.
We say a model \emph{fails} when it makes a prediction that incurs error greater than a threshold:
\begin{equation}
    \ell\left(f\left(x\right), y\right) > \delta
    \label{eq:error_thresh}
 \end{equation}
Here, we introduce some amount of error $\delta$ that the model can make without the prediction being a failure.
This definition both provides a template that formalizes model failure and captures the intuition that a prediction with non-zero error does not necessarily imply that it is a failure.

To make this definition of failure concrete, a practitioner needs to provide an appropriate error function and threshold; both of which are application dependent.
As an example, in a medical setting a clinician may make use of a regression model that predicts the life expectancy of patients in order to determine courses of treatment.
The clinician may get a patient that the model predicts a life expectancy of 80 years, when in reality the patient will actually live 80.5 years.
Strictly speaking, the model has indeed made an error.
If we employ the squared error loss function in \eqref{eq:squared_error} we can quantify the error and report it as 0.25.
Without context of the application of the model, it is hard to reason if this error is significant: Did the model fail with a squared error of 0.25?

In practice, clinicians likely have some level of precision that will practically affect how they will treat patients.
As an example, we may find that clinicians make the same treatment decisions for patients whose life expectancies differ by one year
\footnote{We recognize that this is likely too simple of an assumption, but opt for simplicity for sake of illustration.  More complex definitions of errors can still be captured by \ref{eq:error_thresh}.  In the next section we discuss this point further.}.
We can formalize this definition of failure with the following inequality:
\begin{equation}
    \left(f\left(x\right) - y\right)^2 > 1
    \label{eq:clinician_sq_error}
 \end{equation}
Here, if a model predicts a life expectancy of 80.5 years when the patient will live for 80, it would not be considered an error.
However, if the model predicts 20 years, by \eqref{eq:clinician_sq_error} the model has indeed made an error, because the clinician would have treated the patient differently if given a more accurate prediction.
While this may seem like a simple addition to the formal concepts introduced thus far, it actually highlights an important difference between formal analysis of ML models and understanding of their failure in practice: What constitutes a failure of an ML model is dependent on how the ML model is used.
We will use this definition as a basis for discussing failures going forward.

\examplebox{Running Example: Autonomous Vehicle with Visual Obstacle Avoidance}{Failure in object detection is often defined in terms of cases where the detector makes a prediction that does not correspond to an object in the image (known as \emph{false positives}), or in cases where there is an object in an image and the detector failures to make a prediction that accurately detects it (known as \emph{false negatives}).
Given a single image for which a detector makes predictions, failures are commonly counted by first matching predicted detections with ground truth objects in the scene\footnote{Using a matching algorithm such as the Hungarian Method~\cite{kuhn1955hungarian}.}.
Recall that both predictions and ground truth objects in this setting are defined by a bounding box and class label.
If a prediction has both the same class label as a paired ground truth object, and their bounding boxes are similar, the prediction is determined to be an accurate detection of the object (known as a \emph{true positive}).

These criteria reflect both the classification and localization goals of object detection.
To formalize failure then it must consider both of these.
Let $y_c$ and $y_b$ be a ground truth class label and bounding box, respectively.
Let $\hat{y}_c$ and $\hat{y}_b$ be the same for a prediction.
For classification, \emph{zero/one error} is a common way to determine failure:
\begin{equation}
    \mathbb{1}\left(y_c \neq \hat{y}_c \right)
\label{eq:zero-one-error}
\end{equation}
This simply means that if $y_c$ and $\hat{y}_c$ are the same class, \eqref{eq:zero-one-error} is zero.
Otherwise, it is one.
Localization is commonly measured using some metric such as Generalized Intersection over Union (GIoU)~\cite{rezatofighi2019generalized}:
\begin{equation}
    GIoU\left(y_b, \hat{y}_b\right)
\label{eq:GIoU}
\end{equation}
Intuitively, GIoU compares two bounding boxes by the area in which they overlap (union), and the area where they do not (intersection).
Two perfectly overlapping bounding boxes will have a GIoU of 1, while two bounding boxes that do not overlap at all will have a GIoU less than or equal to 0.

To operationalize this concept of model failure, the obstacle avoidance team find through requirements analysis that the autonomous vehicle may fail to avoid an object if the detector fails to classify it correctly.
From this, they define the following failure condition:
\begin{equation}
    \mathbb{1}\left(y_c \neq \hat{y}_c \right) = 1,
\label{eq:class-failure}
\end{equation}
which formalizes failures due to mis-classificaiton made by the detector.
Similarly, the team finds that in order for the autonomous vehicle to safely navigate around an object it must detect it with a GIoU greater than 0.8.
This gives rise to a second failure condition:
\begin{equation}
    GIoU\left(y_b, \hat{y}_c\right) \leq 0.8
\end{equation}
which formalizes a failure in localization.
These two failure conditions can act as the basis for defining when a model fails and subsequent efforts at reducing failures 
\footnote{In practice, there are many more ways a detector can fail.
Of particular note here is the case where a model fails to detect an object altogether (e.g. a ground truth bounding box has no matching prediction).
This should be included as an additional failure condition that can guide the team to reason about how the detector can fail.
}.
}

\subsection{Machine Learning Operations (MLOps)}
In order to operationalize the concepts discussed in this section, developers require practices, procedures, and processes that facilitate the building of ML models.
This is the focus of machine learning operations (MLOps)~\cite{kreuzberger2023machine}, a form of development operations (DevOps)~\cite{ebert2016devops} for ML software.
MLOps describes a workflow from initial software product initiation through deploying a model.
While many of the subtasks and tooling that is important in DevOps can be applied to MLOps~\footnote{For instance, continuous integration/continuous deployment (CI/CD) tools are often vital in both standard DevOps, as well as quickly iterating on ML models in MLOps.} there are a few key tasks in MLOps that make it unique.
A few of these that are important to failure include:

\textbf{Requirements Analysis} Defining requirements is key to any software system.
However, ML models pose unique challenges not present in other software~\cite{pei2022requirements}.
Of particular relevance to failure are \emph{functional requirements}, which are specifications of what a model should output when given certain inputs.
By the very nature of ML, it is often challenging to define all the kinds of inputs that a model will be exposed to during deployment, as well as the desired model outputs.
Further, how developers take the intuition of what functional requirements should be and turn them into specific formal definitions, such as in the form of failure cases as defined in~\eqref{eq:error_thresh}, may not be obvious.
As such, discovery and specification of requirements for ML models remains an open research topic.

\textbf{Experiment-Driven Model Development} In order to build ML models, developers go through an iterative, experiment-driven process that involves many related tasks.
First, data collection is performed to collect a both training data as well as data used to evaluate models.
Models are then trained and evaluated to determine if they are suitable for deployment.
In practice, this process can involve many more fine-grained tasks, but the basic data collection/training/evaluation process is common amongst all MLOps pipelines.
The results of the evaluation may either give developers enough evidence that a model is suitable for deployment, or it may reinitiate previous steps, ranging from re-training of the model using different design decisions or to re-examining the requirements produced during the analysis phase.
We go more into more detail for each of the data collection, training, and evaluation stages of this pipeline throughout the remainder of this guide.

\textbf{Monitoring} Once models are through the development phase and deployed, they are often monitored for their performance while being used in their deployment environment.
If monitors are implemented, they can be used to flag when ML models fail, and the larger software system around the ML model can determine what appropriate action should be performed.
Developers can also use the outputs of monitors to determine if it is necessary to go back to earlier stages of the MLOps process.
Properly designed monitors can detect when certain performance requirements are not met, such as throughput of model predictions.
In addition, developers can implement functionality where users of the ML model provide feedback at regular intervals and a monitor can flag when a model is unacceptably inaccurate according to users.
However, if users cannot be or are not prompted for their feedback, ground truth labels are commonly not available during deployment and it is difficult to monitor ML models for failures in prediction accuracy.
Indeed, a model may make incorrect predictions that cause a failure, but without ground truth to compare to, a monitor may not be able to detect it.
In subsequent sections we discuss some ways in which a monitor can assess ML model behavior to determine if there is a risk of it failing during deployment.

Next, we use the concepts from this section to discuss the failures in model reliability.
\section{Reliability}
\label{sec:reliability}
\emph{Reliability} is a core concept that spans engineering disciplines, even spawning its own sub-discipline within system engineering~\cite{o2012practical}. 
While different engineering disciplines focus on building different technologies, their definitions of reliability are largely consistent.
For example, the IEEE Standard Computer Dictionary defines reliability as: ``The ability of a system or component to function under stated conditions for a specified period of time.''~\cite{geraci1991ieee}.
This definition has since been adopted by subsequent IEEE standards for cyber systems as well as by the National Institute of Standards and Technology~\cite{ross2021nist}.
The term reliability is also used in the ML community, but inconsistently.
Sometimes it is used interchangeably with model calibration~\cite{guo2017calibration}.
Other times, it is used to describe a property of the process in which one builds an ML model or a system that contains an ML model~\cite{breck2017ml}.
Yet other times, it is used in cases that (perhaps) more accurately describe situations more aligned with the concept of robustness~\cite{lipton2023reliable} (more on this later).
We seek to bring together the common engineering definition of reliability and the common formalisms that ML models are built upon in order to frame potential issues of failures in a way that developers can reason about.

Towards this goal, we begin by deconstructing the IEEE definition of reliability.
In that definition, there are four critical considerations: 1) The ``system or component'' 2) the concept of ``functioning'' 3) ``stated conditions'' in which the system or component should function 4) a ``time period'' in which the system or component should function.
In ML, the system or component can be seen as the model $f$.
In the previous section, we defined what it means for a model to ``function'' by defining failure in \eqref{eq:error_thresh}.
To define the last two points, we look to the foundations we reviewed in the prior section by considering the data distribution during deployment, which we will denote $\tilde{P}\left(x,y\right)$.
If we make the ID assumption, then we assume that $\tilde{P}\left(x,y\right) = P\left(x,y\right)$, or that the distribution that generates deployment data is the same as the one that generated training data.
This implies that the data-generating distribution both during training and during deployment is \emph{static}, that is, it does not change over time.
Given this, we say that the stated ``stated conditions'' are defined by $P\left(x,y\right)$: Any data that is generated by $P\left(x,y\right)$ is within the conditions in which we can say a model is reliable.

By definition, we assume that any $(x,y)$ pair generated by $\tilde{P}\left(x,y\right)$ is independent of other factors, including any other pair that came before or after.
We can interpret this, and the fact that $\tilde{P}\left(x,y\right)$ is static, as defining the ``period of time'' in the definition of reliability for ML models: We can consider ML reliability at the granularity of an individual instance and the prediction that is made by a model.
No matter how long a time period in which we run an ML model, if we make the ID assumption, both the model and the conditions in which the deployment environment generates data remain unchanged.
As such, it is reasonable to define the time period in the definition of reliability to individual inputs and outputs of an ML model.

To summarize, the ID assumption provides a definition of reliability that is based on the fundamentals that underlie much of the practice in building ML models.
Reliability boils down to the ability of a model to avoid failure when given data generated by the same distribution that generated training data ($P\left(x,y\right)$).
This implies that the central challenge in reliability is understanding $P\left(x,y\right)$.
In many applications, the domain of possible instances and their relationship to labels is complex, making reliability a non-trivial property to ensure in ML models.
In the remaining subsections, we will focus on three important topics that are critical in both building reliable ML models, and reasoning about the reliability of ML models.


\subsection{Training Data Collection}
One of the most influential steps in building reliable ML models is collecting high-quality training data.
Intuitively, training data should contain as many and as diverse samples from $\tilde{P}\left(x,y\right)$ as can be obtained
\footnote{In some cases, theoretic results in sample complexity can more specifically characterize how much and what kind of training data will lead to reliable models, but these results often do not give specific insight for practitioners when collecting training data.}.
However, in practice, collecting instances and labels that cover all representative scenarios in which a model can observe during deployment is difficult.
It is often not clear a priori how many samples are needed to train a model or how to reason about the breadth of scenarios that need to be represented in the training data.

For this reason it is important to consider \emph{strategies} for collecting training data, so that there are deliberate steps taken to obtain training data that will lead to reliable models.
Generally speaking, there are two steps to collect training data: Collecting instances $x$, and then gathering supervision on instances in the form of labels $y$.
Different strategies dictate how to collect instances and which of them to label, and often depend on the practical limitations imposed on data collection for a given problem.
Below we discuss three main categories of strategies.

\subsubsection{Passive Data Collection}
For many applications, developers have no direct control over what data is collected.
When building a model that can predict life expectancy of patients, clinicians may only be able to obtain data from patients that come to their clinics.
Even if they had insight into the kinds of patients they want to collect data for, they may not be able to collect data on those patients if they do not come in for care.

In this way, data collection is \emph{passive}: Instances used to train a model are gathered from an environment without any explicit action from a developer to guide what instances they use.
Passively collecting instances in bulk has proven successful strategy in ML model building, especially for complex problems and when there are publicly available sources of instances to be collected from.
Indeed, simply adding labeled instances to training sets without much consideration of which instances are collected has shown to significantly increase model generalization~\cite{russakovsky2015scaling}.
Many of the recent advances in ML owe much of their success to passive data collection strategies at web-scale~\cite{OpenAIData, ravi2024sam, touvron2023llama}.

The main drawback of passive data instance collection is that it does not provide a means to rectify gaps in training data.
Of particular note, passive data collection may exhibit \emph{sampling bias}.
Again, consider the clinical model setting.
Developers may only be able to collect data passively at few geographically similar locations, even though the intent for the model trained on that data is to be used in a broader set of clinics.
Such data may be biased in that it does not represent the geographical diversity of the deployment setting, which in turn can result in a model that fails to generalize to patients in clinics not represented in the training data
\footnote{This example highlights one the most publicized common outcomes of biased ML models: disproportionate harm done to underrepresented human populations~\cite{kay2015unequal, buolamwini2018gender, simonite2018when}.  However, biased training data can cause disproportionate failures in subpopulations of the target distribution in ways unrelated to social bias. See the ``Running Example'' for sampling bias without strong ties to social bias.}.
Here, while the intent is to collect data from the deployment distribution (i.e. $P\left(x,y\right) = \tilde{P}\left(x,y\right)$), due to sampling bias the training data does not actually reflect the deployment distribution (i.e. $P\left(x,y\right) \neq \tilde{P}\left(x,y\right)$).
If developers do not have control over how data is passively collected, they have no means to rectify any biases introduced by the data collection strategy.

Perhaps the most common way to rectify issues in bias is to resample instances from passively collected data to better match the distribution of data during deployment.
One typically either under-samples data from overrepresented subpopulations by removing instances, or over-samples data from underrepresented subpopulations by duplicating instances.
Under/Over-sampling techniques range from simple sampling at random (i.e. randomly resample already collected data to remove/duplicate) to more sophisticated synthetic approaches such as SMOTE~\cite{chawla2002smote}.
While resampling can often be effective, rectifying issues in passively collected data remains an open research topic that continues to motivate the development new and more sophisticated techniques~\cite{hopkins2023designing, hort2024bias}.

\subsubsection{Targeted Data Collection}
The most straight-forward way of preventing bias in passive data collection is to understand the domain in which the model will be used, and collect training data that reflects the same conditions as those during model deployment.
This can be viewed as a \emph{targeted} data collection strategy that can make up for underrepresented populations in training data.
The basic intuitive idea behind targeted data collection is that a developer defines a complete set of settings in which a model will be deployed, and strategizes to collect training data for all of them.
In this way, the training data represents a \emph{coverage} of model deployment settings.

For many applications, defining all the different deployment scenarios for a model is challenging.
Developers would need to define all the different characteristics of the environment that influence the relationship between instances and labels, and then collect data for each unique setting.
For sufficiently complex problems, this is infeasible.
In reality, targeted data collection is used in tandem with some method of discovering \emph{failure modes} of models, where initial attempts at data collection are then supplemented with a process of finding scenarios where training data is lacking.
In the running clinical example, a model trained on passively collected data may disproportionately fail in clinics in certain geographic areas.
Developers could attempt to rectify this issue by collecting data in these clinics, and retrain the model with this new training data.

The need to find and rectify failure modes motivates two important considerations when deploying ML models.
First, it is important to regularly monitor and audit ML models for failure during deployment.
Without monitoring, it is difficult to determine if the deployment environment regularly produces data that represents an undiscovered failure mode.
ML model monitoring is a developing discipline in the engineering of ML-enabled systems with an emerging set of frameworks and tools to support practice\cite{eck2022monitoring,pykes2023aguide,evidently2024model} 
Second, it is critical to develop rigorous, scenario-driven test and evaluation procedures that can uncover important failure modes before deployment.
We will discuss monitoring more in Sec. \ref{sec:reliabilityUQ} and evaluation in Sec. \ref{sec:empiricallyevaluating}, but note here that targeted data collection often relies on one or both to find failure modes that provide insight into what data to collect.

\subsubsection{Active Data Collection}
\label{sec:activedatacollection}
Most of the discussion on data collection thus far has focused on collecting instances.
In the same way one can passively collect instances or target specific scenarios, the same can be true for labels.
Clinicians may provide diagnoses in the natural course of patient care that can be turned into passively collected labels for a model that predicts disease.
If specific diseases are of interest, one could target clinics that specialize in those diseases to have a more targeted label collection.
However, it is often the case that instances are easier to naturally obtain than labels, as labels often require some level human subject-matter expertise to obtain.
If clinicians need to be employed to label patients with disease outside the normal course of their work, developers would likely have to pay clinicians for their time and domain knowledge to obtain labels, which can be prohibitively costly.

This is the central motivation behind \emph{active} data collection, more often called \emph{active learning}\cite{settles2009active,ren2021survey}.
In active learning the main assumption is that while instances are relatively cheap to obtain, labels are costly, and thus strategies should only label instances if they will lead to better performant models
\footnote{Though rarer in practice, there are applications of ML where obtaining instances is also costly.  For these applications, methods for instance selection have been developed that rely on low-cost information about instances that can guide active learning techniques\cite{magee2016spatial}.}.
The most common form of active learning is \emph{pool-based} active learning where it is assumed that there exists a large pool of unlabeled instances for which an active learning algorithm is tasked to find a subset to label.
Most pool-based active learning methods work as follows.
First, a small, random set of instances are labeled and used as the training set to learn a model.
Then a heuristic is used to score each of the remaining unlabeled instances in the pool for how beneficial they would be to learning if they were labeled.
A subset of these are selected based on their scores, given to a labeler to label, and then added to the pool of labeled training data.
This cycle begins again with training a model on the updated set of training data, and iteratively proceeds until either the model achieves some desired level of performance, or some budget of labeling is reached.

The key consideration for active learning is the heuristic that scores instances.
Commonly, instances are scored in relation to the model that is retrained each iteration of active learning.
For instance, \emph{uncertainty sampling} active learning scores unlabeled instances based on a model's confidence in its prediction
\footnote{We have not introduced the concept of ``confidence'' or ``uncertainty'' in prediction yet, so for right now assume it is a number a model produces when it makes a prediction that indicates how likely the model is to predict the correct label for an instance.  We discuss prediction uncertainty more in Sec. \ref{sec:reliabilityUQ}.}.
Instances for which the current iteration of the model is not confident about are given high scores.
High scoring instances highlight where the model may likely make an error.
Thus, labeling them and introducing them into the training set can allow the model to learn from the correct labels, increase its confidence in these instances, and become less likely to make similar errors.
By comparison, high-confidence instances may not lead to a substantial increase in model performance if labeled and trained on, as the model likely already can already predict their correct label.
Other heuristics range form being based on information theoretic principles~\cite{hwa2001minimizing} to reducing expected risk~\cite{roy2001toward} to analyzing the geometry of how instances are represented\cite{sener2018active}. 

While active learning can result in a more focused, efficient data collection that reduces labeling cost, it can introduce a sampling bias.
A truly unbiased sampling from $P(x,y)$ does not consider model performance, and training on actively collected data can lead to a reduction in reliability when compared to models trained on unbiased data.
Recent work has studied bias introduced by active learning, reasoned about when it can harm reliability, and proposed methods for reducing it~\cite{farquhar2021statistical}.

\subsubsection{Other Data Collection Considerations}
While employing proper data collection strategies is important, it is but one of the many steps to ensure training data matches the deployment environment of an ML model.
In practice, there is a considerable, manual effort required to collect a training set that will allow for a reliable ML model to be learned.
This section on training data collection is by no means comprehensive.
However, we will note two final important considerations when collecting data for training ML models.
First, regardless of collection strategy, training data sets must often be \emph{curated} to remove instances or labels that will reduce the reliability of a model.
Clinicians can enter the wrong information into electronic health records, and sometimes make the wrong diagnosis.
Where possible, such erroneous or noisy data should be removed from training data, and best practices for doing so are beginning to emerge~\cite{bhardwaj2024machine}.

Second, it has become common practice to use \emph{data augmentation} to increase the effective number of labeled instances that can be used during training.
As an example, since the advent of modern deep learning~\cite{krizhevsky2012imagenet} so called ``label-preserving'' transformations of data have been employed.
The intuition is that you can artificially create new instances from observed instances by changing them in some non-trivial way that does not change the label.
When building a model to classify the species of bird from an image, it does not matter if the image is rotated, cropped, translated, resized or blurred; the species of the bird in the image remains the same.
Using these transformations to create new labeled instances can result in models that can classify birds in a wider variety of conditions (e.g. blurry images or images of birds farther away) that can occur in the deployment environment.
For this reason, well-planned, domain-and-task-specific augmentations can sometimes reduce the data collection burden for training an ML model.

\examplebox{Running Example: Autonomous Vehicle with Visual Obstacle Avoidance}{To train the object detector for obstacle avoidance, the development team may employ passive data collection by manually driving the vehicle, equipped with a camera, around their local roads.
The frames of the videos collected from the camera can be labeled for objects needed to be avoided by a navigation system.
The team may spend months driving, capturing instances, labeling them, and then using the labeled instances to train an object detector.
At some point, the team may decide to stop collecting data as they find that the model is not failing regularly and collecting more data does not result in improvements in the model.

After a period, the team finds that during regular testing on the local roads, the model is failing at an increased rate.
After looking at the training data they realize that the training data they collected was from April to August, and now it is January, and in their part of the world a new environmental factor has emerged: Snow.
Because they collected visual data during warm months, they failed to capture the subpopulation of instances in the deployment distribution where snow is visually present.
They devise a targeted data collection strategy to drive and collect instances every time it snows.
To save on labeling costs, the team employ an active learning to select the most informative instances in the pool of snowy images to label.
}

\subsection{Empirically Evaluating ML Models for Reliability}
\label{sec:empiricallyevaluating}
Data collection, and indeed all efforts taken to build models, rely on some way to determine that a model is reliable after training.
In Sec. \ref{sec:generalization} we discussed why theory can give us some understanding of the reliability for broad classes of learning problems under specific conditions, but it often does not give assurance for specific models trained for specific applications.
Where theory cannot provide insight, \emph{empirical evaluation} is commonly employed.
Empirical evaluation of ML models takes the form of experiments that can be used to quantify model performance on real data.
In this way, empirical evaluation shows how a model would perform in the field on specific examples.

In the remainder of this section we lay out a framework for empirical evaluation based on unique \emph{test cases}.
We will argue that breaking evaluation into individual tests provides more insight into the reliability of models than evaluating models for more broad notions of their ability to generalize.
We aim not to provide the definitive method for evaluating ML models, but to offer a starting point for practitioners to develop suites of tests that give evidence of the reliability of models under test.

To begin, consider again the most common way of defining what it means for a model to be optimal from \ref{sec:generalization}: To minimize expected error, that is, the average error the model makes over the entire domain of instances and labels.
Given this, a good starting point for measuring reliability of a model is to directly compute \eqref{eq:risk}, as expected error measures generalization of the model to $P(x,y)$.
Unfortunately, it is impossible to compute \eqref{eq:risk} exactly in most non-trivial applications of ML, as it is impossible to collect all possible instances and their corresponding labels.

Instead, common empirical evaluations approximate expected risk by using a finite sample of \emph{test data} $\mathcal{D}_{test} = \left\{\left(x_1,y_1\right), \left(x_2,y_2\right),..., \left(x_m,y_m\right)\right\}$ collected for the purpose of evaluating a model.
To measure reliability, test data is drawn from $P(x,y)$, the same distribution training data is intended to be drawn from.
However, test data is said to be \emph{held out} from training data in that it consists of labeled instances not present in the training data.
In this way, evaluations can measure generalization to data the model wasn't trained on, which more closely matches deployment where the model will observe instances it was not given during training.
Equipped with test data, one can approximate \eqref{eq:risk} by computing \eqref{eq:empirical_risk} with test data instead of training data.

Though seemingly simple, performing evaluations on held out test data in this way (or very similarly) is the cornerstone of assessing ML model reliability.
Most developers' first true assessment of an ML model is done by assessing its ability to generalize to a test set.
While principled, computing a standard notion of expected error on a large collection of labeled data drawn from $P(x,y)$ often does not give developers specific insight into an ML model's performance.
Questions like: ``Is my model reliable enough to deploy?'', ``Are there specific failure modes in my model that I need to fix?'', and ``Is my model appropriate for how it will be used during deployment?'' are difficult to answer from a single test.
To answer these, developers must develop more targeted evaluations, which represent different test cases that provide evidence of the reliability of a model in different contexts.
Test case evaluations can then be combined into a \emph{suite} of tests that more comprehensively evaluates a model.




\subsubsection{The Anatomy of a Test Case}


In order to develop targeted test cases, developers can frame evaluation as an abstraction of the main concepts used when approximating expected error.
To do this, a test case has four components, each of which can be chosen by the designer of the model evaluation
\footnote{A suite of tests will have multiple test cases.  Thus, each of the components can be unique to an individual test case.  Notationally, each component for the $i$th test case could have its own subscript to make this explicit (e.g. $\ell_i$, $m_i$, etc.).  To simplify notation we omit this subscript unless it is needed to differentiate test cases.}:
\begin{enumerate}
    \item An \emph{evaluation metric} $\ell$, defined in the same way as in Sec \ref{sec:formalism_supervised_learning} and in our definition of failure in \eqref{eq:error_thresh}.
    As before, $\ell$ measures the error a model makes when making a prediction on $x$ relative to ground truth $y$ for a labeled instance $\left(x,y\right)$.
    \item A \emph{test set} of $m$ labeled instances $\mathcal{D}_{test} = \left\{\left(x_1,y_1\right), \left(x_2,y_2\right),..., \left(x_{m},y_{m}\right)\right\}$ for which the model will be evaluated against for this test case.
    \item An \emph{aggregation function} $A$ to combines errors over the test set into a single number that indicates a total degree of error.
    \item An optional \emph{failure threshold} $\delta$, again as in our definition of failure \eqref{eq:error_thresh}, which is used to determine if the error made by the model constitutes a failure.
\end{enumerate}
Together, a test case for a model $f$ takes the form:
\begin{equation}\label{eq:testcase1}
    \overset{m}{\underset{i=1}{A}}\left(\ell(f(x_i),y_i)\right)\left[> \delta\right]
\end{equation}
Here, the notation is meant to convey that the evaluation metric computes the error $f$ makes for each of the $m$ labeled test instances.
The errors over each labeled instance are then combined using an aggregation function to produce a single number, which is then compared to a threshold, if provided.
To recover the standard, single-test evaluation setting, $\ell$ would be the error function used during training, the test set would be all held-out test data collected, the aggregation function would be expectation, and there would be no threshold.

The introduction of the threshold allows for specific conditions that define model failure, as opposed to measuring error and leaving it up to the developers to interpret it.
In \eqref{eq:testcase1}, the aggregation of error over all labeled instances in a test set can be compared against a threshold to determine failure.
Note that this differs from our previous definition of failure in \eqref{eq:error_thresh}, where a failure is defined over a single labeled instance.
To reflect this definition of failure, developers can also develop test cases of the form:
\begin{equation}\label{eq:testcase2}
    \overset{m}{\underset{i=1}{A}}\left(\ell(f(x_i),y_i)> \delta\right)
\end{equation}
Here, the aggregation function combines the failures of the model over individual labeled instances.
Later, we will compare \eqref{eq:testcase1} versus \eqref{eq:testcase2}, and scenarios where one is more appropriate than the other.




\subsubsection{Test Case Design Decision Points}
\label{sec:testcasedecisionpoints}
Abstracting empirical evaluation in this way allows for developers to design a suite of tests that can evaluate models under different conditions and contexts.
In order to do so, developers must consider what settings they wish to focus their tests on, and choose the evaluation metrics, test case sets, aggregation functions, and failure thresholds that best measure reliability for those settings.
Each choice affects what is being evaluated, how it is measured, and how it can be interpreted.
Below we discuss test case components and some important considerations when choosing them.


\textbf{Evaluation Metrics} The evaluation metric in a test defines how error is measured between a model's prediction and a ground truth label.
Often in supervised learning, error functions measure some notion of \emph{accuracy} for a given task.
As an example, squared error as defined in \eqref{eq:squared_error} is one of the most common choices for both the loss function used during training and the evaluation metric for regression models.
It can be seen as a measure of accuracy as it increases as a function of the difference between a prediction and a true label.
In almost all applications of regression, however, labels have units, and the metrics to define error in predictions of those units should reflect natural interpretations of them.
Consider, instead, \emph{absolute error}:
\begin{equation}
    \ell\left(f(x), y\right) = |f(x) - y|
\end{equation}
Here, $|\cdot|$ indicates the absolute value, making the absolute error metric the unsigned difference between prediction and label.
If a regression model is predicting units such as time, length, distance, weight, or other real-valued units pertaining to properties of the physical world, developers could arguably place error more easily in the context of how it effects an application domain if they use absolute error as a metric.
If a model is used to predict patient life expectancy, an absolute error can be easily interpreted as how many years the model was wrong by, which is perhaps more easy to reason about in a practical sense than its square.

This example highlights an important principle when choosing evaluation metrics: They should reflect some notion of error that most easily allows test cases to be interpreted as failures in practice.
The difference between absolute and squared error may seem trivial, as you can easily recover absolute error from squared error, but in other cases metric choice may more seriously affect interpretability of evaluation results.
For classification problems, the most common accuracy metric is \emph{zero-one error}, where a prediction incurs an error of one if a predicted class and class label differ, or zero if they are the same.
In many applications, it is important to consider what the true and predicted classes are in a misclassification, not just if the model misclassifies an instance.
A classic example is in medical diagnosis.
A model that makes a \emph{false positive} if it predicts that a patient has a disease when they in fact they do not have that disease.
This can lead to much different subsequent treatment errors than a \emph{false negative} would, where a patient positive with a disease is predicted to not have it.
For this reason, metrics such as precision, recall, specificity, and sensitivity are sometimes used to distinguish the types of errors a classifier makes, not just whether a prediction and label are the same.

\textbf{Test Sets} In a standard evaluation setting, all test data are used to compute a single value of error.
Often, more specific subgroups within all collected test data can be identified to provide insight into how a model performs in unique cases.
When errors for all test data are aggregated, subpopulations that represent edge cases, high-risk scenarios, known failure modes, and other important contexts can be hidden if model reliability for the rest of the population differs significantly.
A classifier may achieve a high overall test  accuracy, but for a high-risk subpopulation that is only represented in a small subset of the test data, the model could be significantly less accurate.
The only way to identify such reliability differences is to create test cases with test sets reflective of targeted subpopulations.

Much like in training data collection, collecting all relevant data to provide a comprehensive coverage of important test cases is a challenge.
Methods for algorithmically selecting subpopulations of data for test case coverage range from feature space based methods~\cite{mani2019coverage} to inspection of model properties~\cite{sun2019structural}.
In many cases, though, subpopulation discovery for test coverage is a much more manual task.
Important subpopulations can be identified by domain experts, through failure modes found during deployment, or through careful enumeration of important latent characteristics of problems from which subpopulations are defined.
It is largely up to developers to determine which subpopulations are worth special consideration for their own test cases.

\textbf{Aggregation Functions} In almost all cases, when designing tests of the form in \eqref{eq:testcase1}, a simple unweighted average as in \eqref{eq:empirical_risk} is used.
The intuition is that an average makes few assumptions about model or data and each labeled instance in the test set contributes equally to the aggregated error.
In some rare cases, other aggregation functions can be used.
For instance, it is often desirable to determine \emph{worst-case} performance of a model on a test set, and instead the aggregation function could output the maximum error a model makes on a labeled instance.
This way, model evaluators can determine the most extreme case of potential failure a model can make on a subpopulation.

Though powerful, tests of the form in \eqref{eq:testcase2} are considerably less common, and thus it is less clear what an appropriate aggregation function is.
If evaluators would like to know how many times a model fails on a subpopulation, an aggregation can simply count how many times the model makes a prediction above the failure threshold.
If the test set represents a subpopulation where failure is catastrophic, model evaluators may want to know if the model fails on any of the test data.
For this, the aggregation function can perform a logical ``and'' that reports ``true'' if the model makes any failure and ``false'' if it makes none.

\textbf{Failure Thresholds}
Failure thresholds determine whenever a test (as in \eqref{eq:testcase1}) or a prediction (as in \eqref{eq:testcase2}) is a failure of a model under test.
Thresholds should be set to reflect some knowledge of the usage of the model, and can be influenced by system performance when a model is used in a larger ML-enabled system, human performance when a model is used to support human decision-making, knowledge of the environment, or other context for how a model is used.
As a result, failure thresholds can often be derived from \emph{functional requirements}.
Requirements engineering, the process in which requirements are generated, for machine learning is still in its infancy, and recent work has highlighted the difficulties ML presents~\cite{vogelsang2019requirements, villamizar2021requirements, pei2022requirements}.
Nevertheless, there are a number of sources for which functional requirements and thus thresholds can be derived.

When ML models are used in a larger system, system modeling approaches such as \emph{model-based system engineering}~\cite{wymore2018model} can be used to understand how an ML model's prediction influences other system components.
Analysis can be performed, and system component interactions can be modeled using languages such as SysML~\cite{friedenthal2014practical} or AADL~\cite{feiler2006architecture} to determine what kind and how much error a model needs to make in order to cause a system level failure.
Even if formal modeling of a system is not done, \emph{system-level testing} may provide data that shows relationships between the amount of model error and system level failure that can guide the development of thresholds.
When models are used to support human decision-making, end user studies can be performed to understand the decision processes humans make when considering model predictions.
If coupled with understanding of the risks associated with user error, thresholds may be derived so that empirical tests are designed that highlight when a model error induces a human error.
Finally, when neither system nor end-user information is available, analysis of the application domain may uncover risks for which ML modelers may determine conservative estimates of thresholds.

Understanding the relationship between error and failure in an application domain is often difficult at the individual prediction level.
Alternatively, considering failure as an aggregate over a subpopulation as in \eqref{eq:testcase1} can potentially give a coarser measurement of failure that is easier to develop tests for.
Specifically, if error is aggregated using an average that is interpreted as expectation, it can be modeled in the framework of \emph{expected utility}.
Expected utility of model predictions can be used to reason about the benefit of utilizing a model in a domain over alternatives.
As an example, if we naively assume that a clinician will always blindly base their treatment of patients on the diagnosis provided by an ML model, we can directly compare how often, on average, a model will make a misdiagnosis versus a clinician that does not use a model.
If the model makes fewer errors on average than a clinician, and that all errors have the same cost, then by one measure, the model is a net benefit on the diagnosis of patients.
A test can be developed with a threshold that a model must have an expected error less than that of the clinician who uses it in order for it to be fielded.

It is important to note that reasoning about errors in expectations can be severely misleading, and the above example is only illustrative of an over-simplfied way of reasoning about failure.
Most attempts in reasoning about expected utility hinge on specifically understanding the costs of incorrect decision-making, which is difficult to do in practice.
For example, a clinician may make more errors in diagnosis on average, but the kinds of errors may actually result in less harm than those made by an ML model. 
Further, comparing ML model performance versus human performance ignores the realities of how humans will use the model in practice.
In most responsible uses of ML for decision-making, ultimate decision authority should still rest with a human, so comparing model versus human performance does not always reflect the reality of model deployment.
Finally, reasoning about error in expectation ignores potential failure modes hidden by aggregating error.
For these reasons, creating failure thresholds for aggregate measure of errors can be one piece of information to help model evaluators understand the reliability of ML models.
However, care should be taken to ensure that additional tests are designed to ensure important additional contexts are evaluated to increase the practical reliability of a model in deployment.

\subsubsection{Optimizing and Satisficing Tests}
\label{sec:OptandSatis}
Equipped with a suite of tests, evaluators have the means to quantify the reliability of an ML model in order to make important decisions about the readiness of a model for deployment.
If instead they only run one test as in the standard expected error setting, the evaluation can only be interpreted very simply: Either the model achieves low enough test error for the evaluators to be comfortable to field it, or it doesn't.
The value of designing a suite of tests instead of relying on a single one is that each test can quantify reliability in different ways, but how should evaluators interpret these tests, some of which indicate failures and others simply provide some aggregated function of error?

In \cite{ng2018machine}, Andrew Ng categorizes tests into \emph{optimizing} and \emph{satisficing} tests.
Satisficing tests are those within our framework with failure thresholds.
These are tests in which a model must pass to be considered for deployment, and can reflect important safety, user, system-level or other known requirements that a model must meet in order to be fielded.
Optimizing tests, on the other hand, have no threshold.
The goal of a model  on these tests then is to achieve the lowest possible error.
In practice, developers likely will train a number of different models during the course of development.
They will likely consider different model classes, training sets, training algorithm hyperparameter settings, and other details that will produce entirely different models.
Optimizing tests can be used to compare two models that pass all satisficing tests in order to choose among them and to track progress as developers iterate on their model designs and training procedures.

Developers can use satisficing tests during development to determine where critical reliability gaps exist.
Explicit steps to remediate these gaps such as targeted data collection or choosing a different class of models can be taken in order to pass the tests required for deployment.
Optimizing tests can be used where failure thresholds are unknown to give developers broader insight into what improvements can be made.
For instance, a model may pass all known satisficing tests, but expected error over all test data may still be high.
Developers may make the determination that the overall accuracy of the model is likely still too low for deployment.
They then can develop new models, using this optimizing test as a measure of progress.
Additionally, this result could lead them to reason about how a model can pass all satisficing tests and still have high overall test error.
This could be an indication that satisficing tests are incomplete, or that the optimizing test contains data representing irrelevant deployment cases that should not be included in test data.

In summary, the test case framework presented here can be used as follows.
First, satisficing tests are developed to evaluate whether model satisfies identified functional requirements.
A satisficing test should include a test set with data relevant to a requirement, a metric that quantifies a measure of performance relevant to a requirement, and a threshold that is used to determine if the model satisfies the requirement.
Then, a single optimizing test can be developed, without a threshold.
This test should provide a more general evaluation of model performance.
The metric here should measure a general, but relevant form of model accuracy, and the test set should encompass all important scenarios the model may encounter during deployment.
During model development, satisficing tests can be used to determine if a model satifies requirements while the optimizing test can be used to break ties between acceptable models and to track overall model improvement.

\subsubsection{Other Model Testing Paradigms}
In this guide, we've focused on case-based testing derived from principles of generalization.
Other testing methodologies exist that can complement the one we outline here.
For instance, the authors of ~\cite{chandrasekaran2023test} discuss a number of alternative model testing paradigms taken from traditional software engineering practices.
Many of these paradigms can alleviate challenges associated with forming specific test cases.
For instance, differential testing~\cite{mckeeman1998differential} does not require labels for tests and instead compares the output of a model under test to another model.
This can be useful when training a new model that is supposed to maintain the performance of an already trusted model.
Combinatorial testing~\cite{kuhn2010practical} provides methods for pseudo-exhaustively searching the space of possible test samples to achieve more complete test coverage.
In ~\cite{chandrasekaran2024leveraging}, the authors propose combinatorial testing method for the setting where test scenarios in the deployment environment of an ML model can be distinguished by a finite number of factors. 

Recently, ML model \emph{red teaming}~\cite{feffer2024red} has become an increasingly popular paradigm for ML model testing.
Here, a tester makes explicit attempts to find an input to an ML model under test that will cause it to fail.
Red teaming is especially useful in cases where there is very little cost for a tester to create valid inputs to a model, such as models that take natural language as input.
Successful red teaming can help model evaluators create targeted tests that quantify found failure modes, which can in turn provide a way for developers to test remediation strategies.

\examplebox{Running Example: Autonomous Vehicle with Visual Obstacle Avoidance}{After an initial object detector is built, the obstacle avoidance model team wants to test their model.
First, they drive their vehicle around their local area to collect new instances not already collected for training
\footnote{Care must be taken in this case so that instances in the test set are indeed \emph{independent} of those in the training set.  One could naively collect all train and test instances at once, and randomly assign each to one of either the train or test sets.  However, because instances are frames from videos, it is possible that consecutive frames can be split between test and training sets causing a dependency between them.  When test data is either not properly held-out or independent of train data, this is known as \emph{test set poisoning}.}.
The team labels these with bounding boxes for objects the vehicle should avoid.
As a first attempt at evaluation the team decides to use the entire test set and the \emph{mean average precision} ($mAP$) metric \cite{padilla2020survey}, as this is how many object detectors in the computer vision literature are primarily evaluated.

While this does indeed give them a quantitative assessment of their object detector, they have a few concerns.
First, $mAP$ is a metric that measures both localization and classification performance by making few assumptions on how the object detector is used.
It averages precision (correctness of a detection made by the model), over different levels of localization performance (measured by intersection over union (IoU)).
In this way, the model may make many correct predictions when localization performance is low, and few correct predictions when the localization performance is high, and get the same $mAP$ if precision is consistent across localization performance.
The team finds through requirements analysis that in order for the vehicle to safely avoid obstacles that the detector must detect objects with an IoU greater than or equal to 0.8.
This motivates them to choose a different metric: average precision at IoU 0.8 ($AP@0.8$), which only counts a prediction as correct if it localizes an object with an IoU greater than or equal to 0.8.

Even after changing the evaluation metric, the team realizes they have another issue.
They recognize that detecting pedestrians is of critical importance, but by evaluating their model on all objects, they cannot determine if their model is correctly detecting all pedestrians or not.
In response to this observation, they create a test that the object detector must pass in order to be fielded, which has a test set consisting of only labeled instances of pedestrians.
Since they need every pedestrian to be detected, they choose the metric average recall at IoU 0.8 ($AR@0.8$).
This metric is 1 if all objects in an image are detected, regardless of the chosen precision threshold.
Thus, this test case necessitates a failure threshold where a model fails if it achieves an $AR@0.8$ on the pedestrian test set less than 1.

With these two test cases, they have a single optimizing test:
\begin{equation}
AP@0.8(\mathcal{D}_{test}, f)
\end{equation}
and a single satisficing test to ensure all pedestrians are detected in their test data:
\begin{equation}
    AR@0.8(\mathcal{D}_{test}^{pedestrian}, f) \geq 1
\end{equation}
}

\subsection{Model Reliability Self-Assessment and Monitoring During Deployment}
\label{sec:reliabilityUQ}
So far we have discussed ways of improving and reasoning about the reliability of ML models before they are deployed.
Training data collection is a critical step in building reliable ML models.
Empirical testing allows for developers to quantify the reliability of models after they are trained.
Once deployed, though, neither of these can provide a way of ensuring ML models are being used reliably.
This is made explicit in common designs for ML pipelines where data collection happens before evaluation which happens before deployment~\cite{kazmierczak2024MLOps, visengeriyeva2024MLOps}.
However, advanced pipeline designs often include some mechanism for \emph{monitoring}, where assessment of the model during deployment is built into the system, and can re-initiate previous stages in the pipeline when issues are identified.
Proper monitoring can allow an ML-enabled system to react appropriately when a monitor indicates a model is likely to fail.
When models are used to aid human decision-making, monitoring can allow a human end user to reason about courses of actions when failure is likely.
In both cases, monitoring can flag potential issues, so developers can then improve the reliability of models.

Commonly, monitoring software systems involves tracking and checking the inputs and outputs of subsystems or components~\cite{pietrantuono2010online}.
Monitoring ML models for their reliability in this way presents a unique challenge because seemingly valid, in-specification instances can cause a model to fail, and creating test cases for all possible failure modes is often practically impossible.
In most cases, the only definitive way of monitoring for model reliability is to check model predictions against ground truth labels, and if ground truth is available during deployment, there is no real need to use ML at all.

This makes monitoring ML models for reliability during deployment seem hopeless.
However, there exists an active field of machine learning dedicated to model \emph{self-assessment}~\cite{canal2024decision}.
Here, models not only output predictions for each instance, but also some quantitative assessment of the model that can be interpreted by humans or monitoring mechanisms.
In the remainder of this section, we will discuss topics in model self-assessment, starting from a common way which models articulate assessments and how practitioners can use assessments in monitoring model reliability during deployment.

\subsubsection{Probabilistic ML Models}
\label{sec:probML}

As discussed in Sec. \ref{sec:generalization}, a major assumption made in the theory of supervised ML is that data during deployment is generated by a distribution $P(x,y)$.
Because learning $P(x,y)$ exactly is often impractical resulting in models that can fail, it is often useful to make further assumptions about the relationship between how data is generated from an environment and a model learned from that data. 
These assumptions can often lead to ways of learning models that express how they make predictions \emph{probabilistically}.
Estimates of probabilities that influence how a model makes predictions can reveal when and how a model is ``uncertain'' about predictions, a key indicator of model reliability
\footnote{Those already familiar with probabilistic ML might find the treatment in this section different than how it often is presented elsewhere.  We opt for a utilitarian treatment of the topic that emphasizes how probabilities can be used, rather than one that starts from first principles.  For an introduction to probabilistic ML that is more grounded in probabilistic foundations, see \cite{murphy2022probabilistic}}. 

Perhaps the most common assumption made in supervised learning is that the data generating process can be decomposed using rules of \emph{conditional probability}.
One such decomposition takes the form $P(x,y) = p(y|x)p(x)$.
Here, the term $p(x)$ is the probability of an instance, and $p(y|x)$ is the probability of a label being assigned to that instance.
This assumes a process where an instance, then its label are generated by an environment.
When making a prediction, a model is \emph{given} the instance $x$, thus to find $P(x,y)$ only $p(y|x)$ needs to be computed.
As a result, the task of learning a model can be simplified to only learning $p(y|x)$, which is often called a \emph{predictive probability}.

Models that learn predictive probabilities often do not output individual predicted labels when given an input, but instead output an estimate of $p(y|x)$.
In this way, models that output predictive probability distributions are more expressive than ones that simply output predicted labels.
If predictions are desired for an application, one can simply turn predictive probabilities into predictions by taking the most probable label (i.e. the $y$ that maximizes $p(y|x)$).
However, predictive probabilities can be used to reason about the uncertainty in predictions, as well.
In classification problems, $p(y|x)$ takes the form of probabilities over each class in $\mathcal{Y}$
\footnote{To illustrate this more clearly, a model that can classify an instance as one of three classes (3-way classification) would output three probabilities, each corresponding to one of the three classes. If a model takes an instance $x$ and outputs $\left[0.7, 0.2, 0.1\right]$ the model has assigned the highest probability to the first class, which can be used as the predicted class for $x$.}.
Class predictive probabilities are often called the \emph{confidence} that the model has in each class.
When a class is assigned a high confidence, the model has estimated that the instance is likely to be a member of that class.
High confidence can be interpreted as the model assessing that its prediction is reliable.
More nuanced assessments of reliability can also be quantified through confidence assessments.
If there are two classes that are assigned similar, high confidences, then the model is uncertain whether correct label is one or the other class.
If all classes are assigned relatively similar confidences, the model estimates the instance may be a member of any class, signalling that its prediction is highly unreliable.

The intuition behind these interpretations can be formalized using metrics that quantify the uncertainty of model predictions.
A system can then monitor these metrics during deployment, and can flag when the model is uncertain.
For instance, the \emph{maximum predictive probability} metric, defined as the highest probability output by the model, can be used to monitor the confidence in predictions.
The \emph{relative predictive probability} metric, defined as the absolute difference between the highest and the second-highest probabilities, highlights whether the model is uncertain between two possible predictions.
\emph{Entropy} takes into account all predictive probabilities:
\begin{equation}
    -\sum_{i=1}^{n}p(y_i|x)\log\left(p(y_i|x)\right),
\end{equation}
which is maximized when all classes are assigned the same probability value.

Fortunately for developers who want to monitor their models during deployment, most modern classification models have intuitive ways of interpreting their outputs as predictive probabilities.
Many regression models, on the other hand, often simply output a predicted label with no obvious probabilistic interpretation.
To obtain predictive probabilities in regression, developers often have to choose classes of regression models that take explicitly probabilistic forms.
Because the domain of labels, $\mathcal{Y}$, in regression problems is infinite
\footnote{Recall that in regression, $\mathcal{Y} = \mathbb{R}$, and the set of real numbers is infinite.  Even if a model only outputs predictions over a range of real-values, the domain of labels is still infinite.  If instead, the model outputs a finite set of real values, the problem is then not called regression but \emph{ordinal regression}, and probabilities can be assigned to ordinal labels similarly to how classification models assign probabilities to classes.}
regression models cannot simply output a probability for every label as in classification.
Instead, they typically specify probability distributions that can be used to assign probabilities to labels and used to quantify predictive uncertainty.

Due to the breadth of approaches, we choose not to provide a review of probabilistic regression models in this guide.
However, one illustrative and important common distribution used to express predictive probabilities in regression is the \emph{normal distribution}.
Here, regression models output a \emph{mean} $\mu$ and a \emph{covariance} $\sigma$ when given an instance as input.
Commonly, the mean is interpreted as the prediction of the model, as the mean is the most probable event in a normal distribution.
However, other interpretations of normal distributions lead to estimates of uncertainty.
For instance, the entropy of a normal distribution is defined as a function of the variance: $1/2\log\left(2\pi{e}\sigma^2\right)$\footnote{Here, $\pi$ and $e$ are the mathematical constants for the ratio of a circle's circumference to its diameter and Euler's constant, respectively.}.

\subsubsection{Monitoring Self-Assessments}
Even with an accurate metric for self-assessment, a proper monitoring system must still have some method for determining when a model is unreliable.
Stated another way: How do developers determine how uncertain is too uncertain?
Just as in developing test cases for evaluation, in practice, what to measure, what constitutes an unacceptable risk, and how a system should react when a model is uncertain are all dependent on the application.
Proper functional requirements analysis should be performed to best understand how to best monitor models.
Monitors should involve carefully chosen metrics that measure probabilities of interest, and thresholds for these metrics that reflect what is known about an application as revealed by requirements analysis.

When thresholds are not established a priori, self-assessments can still be monitored using techniques from \emph{anomaly detection}~\cite{chandola2009anomaly} and \emph{change point detection}~\cite{aminikhanghahi2017survey}.
Anomaly detection (specifically, what is called ``unsupervised'' anomaly detection) methods aim to detect rare events when only nominal data is available.
If a baseline for self-assessment metrics can be established using known cases where the model is reliable, these techniques can be used to detect when a model is unreliable.
Change point detection methods perform similarly, but include the concept of time.
They assume that data is generated over time via a probability distribution, and the goal is to detect if and when data begins to be generated by a different distribution.
When a model is deployed, a self-assessment metric can be monitored and a nominal distribution of its values can be established.
A change point detection method can be used to continue to monitor metric values and detect if they suddenly change.
A sudden drop can indicate an event that has affected model reliability.

As part of our discussion on evaluation in Sec. \ref{sec:testcasedecisionpoints}, we discussed how expected utility can be used to reason about the affect of failures.
Probabilistic self-assessment provides some of the basis for reasoning about immediate courses of action during deployment using expected utility.
Specifically, accurate estimates of predictive probability can be used to reason about the relative probability of different labels being applied to an instance during deployment.
When combined with accurate estimates of the cost associated with failing, a monitor can effectively weigh the risks and probability of outcomes resulting from model predictions to determine the best courses of action.

\subsubsection{Model Calibration}
Just as a prediction from a model can be inaccurate, so can a model's estimate of predictive probability.
In principle, in order for estimates of probability distributions to be accurate, they should reflect the true probability of the events they are modeling.
This means that predictive probabilities output by ML models should match the actual probabilities of instances being assigned labels in the environment.
This is the goal of model \emph{calibration}, and is formalized through a \emph{calibration condition}\footnote{Much of the ML literature actually refer to this as a \emph{reliability condition}. We opt instead to call it a calibration condition as to not overload the term ``reliability''.}:
\begin{equation}
\label{eq:calcond}
    \forall_{f(x)}P(y|f(x)) = f(x)
\end{equation}
Here, we assume that our model $f(x)$ is probabilistic and outputs an estimate of $p(y|x)$.
\eqref{eq:calcond} defines what it means for a model to be calibrated: the predicted probabilities output by a model (right hand side) should match the actual probabilities of a label given the output of the model (left hand side).
As an example, consider a probabilistic binary classifier that outputs a single probability of the positive class\footnote{Note that the probability of the negative class can be recovered by computing $1 - f(x)$}.
If the classifier is calibrated, then one could take all instances where the model outputs a probability of 0.7, and the true label of those instances would be the positive class 70\% of the time.
The same would be true for all probabilities output by a calibrated model, not just 0.7.
Defining accuracy of predictive probabilities under this lens of calibration gives rise to the interpretation of predictive probability as confidence.
If a calibrated model predicts a label with high probability, then the model is more likely to be correct in that prediction.
Thus, the predictive probabilities can be seen as a model's self-assessment of how confident it is in being correct.

Because probabilistic regression models output continuous probability distributions instead of assigning probabilities of each class, calibration conditions and resulting evaluating metrics for regression are different than those for classification.
One way of defining a calibration condition for regression models first introduced in ~\cite{gneiting2007probabilistic} is by defining it in terms of \emph{confidence intervals}.
Here, we assume that when given an instance $x$ and a probability $p$, the outputs of a model $f$ can be used in defining an upper and lower bound for a confidence interval \footnote{As an example, any probabilistic regression model $f$ that outputs a mean $\mu$ and a variance $\sigma$ of a normal distribution when given an instance $x$ can produce an upper bound of a confidence interval defined as  $\mu+\sigma\Phi^{-1}(p)$ and a lower bound defined as $\mu-\sigma\Phi^{-1}(p)$, where  $\Phi^{-1}(p) = \sqrt{2}\mathrm{erf}^{-1}\left(p\right)$ and $\mathrm{erf}^{-1}$ is the inverse of the Gauss error function.}.
With these, we can define a calibration condition that states that for all probabilities $p \in [0,1]$ the true label for any instance should be between the upper and lower bounds of the confidence interval $(p*100)\%$ of the time.
As an example, for a calibrated model its 0.7 confidence interval should contain the true label between its upper and lower bound for 70\% of instances\footnote{It is important to note that regression models can be trivially be calibrated by always providing the largest possible confidence interval (i.e. $[-\infty, \infty]$).  Such large intervals are not useful for a monitoring system that needs to reason about the likely labels an instance can take.  This is why probabilistic regression models should not only be calibrated, but \emph{sharp}, which intuitively means that intervals should be as small as possible. In this way, calibration and sharpness are competing properties in probabilistic regression.}.
Metrics that measure uncertainty in probabilistic regression models either directly or indirectly (through use of common parameters) use confidence intervals.

\subsubsection{Learning Calibrated Models}
\label{sec:learningselfass}
While many classes of models naturally express predictive probabilities that can be used in reliability monitoring, some of the most widely used training procedures that have been successful in learning accurate models are known to produce models that are mis-calibrated~\cite{guo2017calibration, leng2024taming}.
This has motivated the development of techniques explicitly aimed at producing better calibrated predictive probabilities.
Broadly, techniques to learn calibrated models fall into one of three categories.
First, some fundamental modeling assumptions tend to lead to better calibrated models than others.
In this way, these assumptions are intended to \emph{inherently} produce calibrated models.
For instance, it has been reported that Bayesian Neural Networks (BNNs) \cite{heek2018well, mosser2022comprehensive} and Deep Ensembles \cite{lakshminarayanan2017simple} tend to be more calibrated than standard neural network modeling paradigms despite having very similar loss functions, training algorithms, and model classes.

Second, there are design choices that can result in better self-assessment capabilities.
Using certain loss functions during training, such as variations of \emph{focal loss}~\cite{mukhoti2020calibrating,ghosh2022adafocal,tao2023dual}, has been shown to produce better calibrated models than those most commonly used.
Similarly, some data augmentation techniques have been shown to improve calibration~\cite{wen2021combining,rao2023studying}\footnote{Interestingly enough, both of these references also show that some augmentations \emph{hurt} the accuracy of predictive probabilities output by models.  This highlights the delicate relationship amongst decisions made during training and the various ways to measure the performance of learned models.}.
Finally, some model classes have been shown to produce more accurate predictive probabilities than others, when otherwise trained in the same manner.
As an example, transformer networks have been shown to produce better predictive probabilities than alternative neural network architectures \cite{minderer2021revisiting}.

The last category is \emph{post-hoc} techniques that are used to improve model self-assessment after a model has been trained.
Post-hoc calibration techniques learn an additional model $g$ that takes the predictive probabilities of a trained model $f$ as input and outputs probabilities that are intended to be better calibrated.
During deployment, $g$ can then be applied to the outputs of $f$ for the purpose of model self-assesment.
Most post-hoc techniques are agnostic to the models they are used on and do not require significant changes to how one designs or trains a model.
For these reasons, post-hoc techniques are an attractive way to improve self-assessment without taking special considerations elsewhere when building models.

Post-hoc calibration techniques for classifiers vary widely in their complexity and performance.
Early techniques are either non-parametric, such as histogram binning~\cite{zadrozny2001obtaining} and isotonic regression~\cite{zadrozny2002transforming}, or had few parameters, such as temperature scaling~\cite{platt1999probabilistic}.
Because these techniques are simple, they require little additional effort to learn and are computationally efficient to use during deployment.
Since then, more sophisticated techniques have been developed.
For example, in response to findings of fundamental issues with early techniques, better-performing post-hoc calibration techniques that combine ideas from early models have been developed~\cite{kumar2019verified, gupta2020distribution}.
Post-hoc calibration of regression models has been less of a focus from the ML research community than classification, but there does exist promising methods specifically for calibrating regression models~\cite{kuleshov2018accurate}

\subsubsection{Evaluating Models for their Calibration}
Rarely do probabilistic models satisfy \eqref{eq:calcond} exactly, nor would it be practical to check if a model is perfectly calibrated for all possible instances.
Instead, practitioners can evaluate model calibration using metrics to measure how far they are from satisfying \eqref{eq:calcond} and a test set in the same way models are evaluated for accuracy.
The most commonly used metric for evaluating classifier calibration is \emph{expected calibration error} (ECE)~\cite{naeini2015obtaining}, which measures the average absolute difference between a models maximum predictive probability, and it's accuracy.
In doing so, ECE measures how well the predictive probability assigned to the most probable class reflects the actual probability of the model predicting the correct class.
Metrics analogous to ECE have also been developed for regression \cite{kuleshov2018accurate}.

ECE is a relatively complex metric that makes a number of assumptions ranging from statistical assumptions that affect the estimation of the metric to practical ones that affect how the metric should be interpreted in practice.
This has lead to alternative calibration metrics to be proposed.
The authors of \cite{nixon2019measuring} propose \emph{adaptive calibration error} (ACE) which addresses issues in the statistical estimation of calibration error and expands ECE to measure calibration beyond the highest predicted probability.
In \cite{kirchenbauer2022your}, the authors propose a framework called \emph{generalized calibration error} (GCE) that can create metrics that measure calibration with respect to how predictive probabilities will be used in practice.

\subsubsection{Other Reliability Self-Assessment Paradigms} 
Predictive probabilities -- when interpreted through the lens of calibration -- can provide a powerful tool for monitoring the reliability of ML models, but they are not the only mechanism in which models can self-assess.
In the \emph{learning to reject}~\cite{bartlett2008classification, yuan2010classification, cortes2016learning, mao2024two} setting, models are trained with the ability to reject an instance instead of make a prediction.
This can be interpreted as the model abstaining from making a prediction due to high uncertainty or estimating that the expected cost of abstaining is lower than the expected cost of making a prediction\footnote{In the latter case, some concept of cost is needed, namely, a cost associated with the model rejecting an instance and a cost for the model failing.  Models then can estimate the probability of failure and weigh that relative to the cost of abstaining or making a prediction.}
The benefit of these models is that they have monitoring capabilities built into them.
All a system would need to do is monitor for if the model outputs a rejection.
The downside is that rejection models can be opaque as to why they may reject an instance.
In practice, it may be beneficial to manually build monitoring rules and strategies for handling uncertainty that are directly informed by what is known about an application domain and would otherwise be difficult to capture in the frameworks provided by learning the reject methods.

There are even alternative probabilistic modeling options to predictive probabilities.
A popular alternative to models that output predictive probabilities are ones that perform \emph{conformal prediction}~\cite{shafer2008tutorial,lei2014distribution, angelopoulos2021uncertainty, ghosh2023improving}.
Instead of a model producing a probability that can be interpreted as a confidence in a prediction, in conformal prediction the model takes a probability $p$ as input and the model outputs a \emph{set} of possible labels than an instance can take with probability $1-p$.
This is useful in situations where model outputs are used to make decisions where multiple alternatives can be reasoned about.
A conformal predictor can provide multiple labels for which one is correct with high confidence.
Additionally, conformal prediction has strong theoretical backing.
Conditions for which a conformal predictor will be correct are well-established, and there are known principled ties to calibration, as well as prediction of confidence intervals~\cite{gupta2020distribution}.

\examplebox{Running Example: Autonomous Vehicle with Visual Obstacle Avoidance}{
Even though the obstacle avoidance model team has built an object detector that has passed a carefully constructed suite of tests, they find that it still occasionally fails on instances they would otherwise expect it to make correct predictions on. 
To better handle model failures during deployment and to more efficiently collect failure cases for the team to triage, they decide to build a monitoring system that can track self-assessments of their model.
They choose to track the highest class confidence (i.e. the maximum predictive probability) output by the model each time the model makes a detection, because they want to catch cases where the object detector is uncertain in its predictions.
Not only will the monitoring system keep a record of uncertain instances it encounters during deployment for the development team to consider later, but it will also be used to determine the vehicle's behavior.
Through proper requirements analysis, the team found that there is an unacceptable risk of colliding with high-consequence objects (people, other cars, etc.) if the object detector has a highest class confidence less than 0.95 for three consecutive frames of the video feed
\footnote{This is a simple example, but note that scenarios and their associated risk here are more complex. For instance, what may be of even higher risk is when a model makes no detection at all rather than making a detection with low class confidence. Different object detectors have different ways of determining when a detection is of such low confidence that it can be ignored.  Developers should consider details like these when reasoning about risk and probability of failure during deployment.}.
If this happens, the monitoring system initiates a ``safe mode'' that slows the vehicle, alerts human passengers of its uncertainty in its environment, and prompts them to take control.

Because the team chose to build their object detector from the YOLO class of models, it expresses class probabilities naturally through the last layer of the classification portion of the neural network.
\footnote{It is important to note that standard YOLO models do not output probabilities for their localization predictions.  If the team wants to track uncertainty in the location of objects, they would need to change either the class of models they use or how they train their model or both.}.
To test the highest class confidence output by the model, they create a new set of test cases that evaluates calibration using ECE
\footnote{Note that ECE measures the calibration of the highest class probability, but it does so over all confidence values of the model. For this case, confidence is used to distinguish between a ``high confidence'' case (greater than 0.95), and a ``low confidence'' case (less than 0.95).  Thus, errors such as the model outputting 0.6 when it should have output 0.1 can be significant errors as measured by ECE, but do not matter in practice, as both would be correctly monitored as low confidence predictions.  Considering this, it might be more appropriate to use a specialized metric derived from GCE that measures errors in the model between the low and high confidence categories.}.
They find that the model has relatively high ECE in some important cases, and find that it is regularly \emph{overconfident} in its predictions.
To remedy this, they collect a calibration set of labeled data and apply the post-hoc calibration technique temperature scaling to their model.
After they do so, they find the confidences output by the model after temperature scaling pass their suite of tests.
}

\subsection{Summary}
In this section we defined reliability as the ability of a model to make correct predictions when given data that is in-distribution, that is, data from the same distribution as the one it was trained on.
Using this definition as a starting point, we reviewed three topics relevant to building, evaluating, and monitoring models for their reliability.
First, we discussed strategies for collecting training data, including the benefits and drawbacks associated with each.
Second, we introduced a framework for test-case-based evaluation of ML models that can allow developers to identify the contexts in which their models are reliable, and contexts where they are not.
Third, we reviewed topics in model self-assessment, which can be used to monitor the reliability of models during deployment.
While these topics are not a comprehensive coverage of all the considerations necessary to build and use reliable ML models, we believe they provide an important basis that can help guide development teams.
In the next section, we will shift focus from reliability to robustness of ML models.
Many of the topics we've discussed in this section will be relevant.
However, we will show why there is an important distinction between the two properties that must be considered when building ML models.


\section{Robustness}
\label{sec:robustness}
Just as reliability has a long history in engineering, so does robustness.
The IEEE defines robustness in software as ``the degree to which a system or component can function correctly in the presence of invalid inputs or stressful environmental conditions''~\cite{ieee_1990}.
The key difference between this definition and the definition of reliability is that the system or component encounters \emph{invalid inputs} or \emph{stressful environmental conditions}.
This can be understood as inputs that are not part of the specification of assumed operating conditions when the system or component was designed and built.
As we discussed with reliability, the main assumption made about the deployment environment that makes ML models generalize is that data encountered during deployment are generated from the same distribution as the training data.
As such, robustness in ML can be defined as a model's ability to avoid failure when it receives data from a different distribution than the data it was trained on.
Formally, ML robustness concerns itself with the case where the training distribution $P(x,y)$ is not equal to the deployment distribution $\tilde{P}(x,y)$ (i.e. $P(x,y) \neq \tilde{P}(x,y)$).
When this occurs, it is often called \emph{distribution shift} and data encountered from a shifted distribution is called \emph{out-of-distribution} (as opposed to \emph{in-distribution}).

Practically, a distribution can shift for a number of reasons.
ML models are often deployed in dynamic environments where conditions can change over time.
Training data collected at one time may be fundamentally different from data generated from the same environment in the future.
Even if models are deployed in relatively static environments, design and development decisions made by developers may result in unhandled shifts in data.
Developers may attempt to deploy a model that was successful in one environment to another without fully appreciating the differences in environments.
Requirements might change, affecting the ground truth of labels and making previously correct predictions now incorrect.
Biased data collection may result in training models for a distribution that is not actually reflective of the deployment environment.
These and many other possible scenarios highlight the need for methods to model distribution shifts that may affect model robustness.

Fortunately, ML researchers and developers alike have been fairly consistent in using definitions of model robustness tied to distribution shifts, which has led to considerable recent focus on the topic.
However, different approaches to robustness, either explicitly or implicitly, make considerably different assumptions about \emph{how} data has shifted.
Upon initial consideration, it may not seem clear why this is necessary; Can we make models that are robust to any kind of change from train to deployment?
Recall again that the foundations we discussed in Sec. ~\ref{sec:generalization} explain how ML models generalize to new data from the training distribution using the ID assumption.
With the ID assumption being broken by out-of-distribution data, theoretical explanations of reliability do not hold for robustness cases, and it is unclear why it should be expected that models generalize. 
Indeed, research has proven that learning a model that generalizes from $P(x,y)$ to $\tilde{P}(x,y)$ when $P(x,y) \neq \tilde{P}(x,y)$ and no assumptions are made about the relationship between the two distributions is \emph{impossible}~\cite{david2010impossibility, lipton2018detecting}
\footnote{This is true even if unlabled instances from $\tilde{P}(x,y)$ are available during training.}
\footnote{In ~\cite{garg2021leveraging} the authors also show that even estimating the accuracy of any model trained on $P(x,y)$ and evaluated on $\tilde{P}(x,y)$ is impossible, so one cannot even anticipate failures after a distribution shift if no assumptions are made.}.
Intuitively, without being able to rely on principles of model generalization in-distribution, models must rely on something else to relate training and deployment data.
Unless a model can make some assumption between the two, any relationship can be true during deployment and it can fail arbitrarily 
\footnote{An alternative way of thinking about why generalized robustness is difficult is through the concept of a \emph{world model}. A world model is a construct in which an ML model can simulate a deployment environment in order to reason about the relationship about instances and labels.  To achieve complete robustness a world model would need to be able to model every possible instance/label pair~\cite{mccarthy_1969_philosophical}, which is practically impossible for many applications.  Recent work has tried to view robustness under the lens of world models and outlined roadmaps for how more powerful world models can lead to more robust ML models~\cite{dalrymple2024towards}.}.

Impossibility results, however, do not mean that achieving robustness in any setting is impossible.
They do motivate the creation of formal assumptions that can be the basis for methods that learn robust ML models.
In the remainder of this section, we will discuss different assumptions in the form of \emph{models of shifts} that can be used to enable model robustness.
We begin by reviewing some formal assumptions, and techniques based on them, that make reasoning about out-of-distribution data and learning robust models possible.
These techniques either \emph{detect} when shifts occur or \emph{adapt} models to shifted distributions.
We end this section by more directly relating robustness to the previous section of reliability by discussing it in the context of data collection, testing, and monitoring.

\subsection{Model Uncertainty}
\label{sec:paramuncertainty}
Perhaps the most obvious place to begin a pursuit in detecting when data has shifted is by using predictive probability.
After all, if an instance given to a model is somehow different from the data it was trained on, we might expect the model to be uncertain about a prediction.
Unfortunately, it has been shown that many modern ML models can produce high confidence predictions on examples that are clearly out of the domain of the training data, such as noise, or images of artificially generated patterns ~\cite{nguyen2015deep}.
This highlights that most standard probabilistic ML models do not explicitly capture the uncertainty caused by lack of relevant training data when computing predictive probability.
Even if techniques are used to calibrate models, the calibration set is typically from the same distribution used to train the model, and are likewise not suited for modeling how predictive probabilities should reflect when data is from a different distribution.

In contrast to standard probabilistic ML models, \emph{Bayesian} models have explicit terms that are used in computing predictive probability that model the relationship between learned models and data.
In most cases, model classes are \emph{parametric}, meaning that a model from a class is uniquely characterized by a set of parameters $\theta$ that determine how it makes predictions
\footnote{For example, linear models that take the form $f(x) = wx + b$, a model is parameterized by its weight $w$ and bias $b$. A common training algorithm will find a setting for $w$ and $b$ that minimizes a loss function with respect to training data.}.
We make this explicit in our notation by re-writing the predictive probability of a model with parameter setting $\theta$ as $p(y|x,\theta)$.
While a standard supervised model would have a fixed parameter setting $\theta$, chosen by a training algorithm, methods that train Bayesian models learn a distribution over parameter settings $p(\theta|\mathcal{D})$ called a \emph{posterior distribution}.
Intuitively, the posterior models the relationship between data and model parameter settings by assigning higher probability to settings that achieve lower training loss.

A single parameter setting can be chosen by selecting the $\theta$ that maximizes the posterior, as it is the most probable model given the training data.
This is known as \emph{maximum a posteriori estimation} of a model.
However, this loses much of the power of having a distribution over parameters.
Alternatively, we can utilize the full posterior distribution by computing predictive probability using the following:
\begin{equation} \label{eq:posteriorpp}
p(y|x) = \int_{\theta \in \Theta}p(y|x,\theta)p(\theta|\mathcal{D})d\theta
\end{equation}
Here, predictive probability is found by integrating over all parameter settings.
One can view \eqref{eq:posteriorpp} as taking the average predictive probability over all parameter settings, but weighing them by their posterior probability.
If models with better fit to training data have higher posterior, they have stronger influence on the final predictive probability $p(y|x)$.
This highlights the power of Bayesian modeling: If many different parameter settings fit well to training data, and are assigned high posterior probability, all of them can be used in making predictions instead of just using one that has particular bias outside of the training set
\footnote{To illustrate this point further consider the following.  A properly trained Bayesian model posterior would assign high probability to all parameter settings that achieve low training error.  As a result, these models should each produce similar predictions for instances similar to training data instances.  But what about instances dissimilar to the training set?  Minimizing training loss does not encourage training algorithms to choose particular models dissimilar to training data, and thus many different models with low training error may produce very different outputs for instances dissimilar to training data.  If only one of these models is chosen, then the training algorithm commits to one of many possible models that can fit well to training data.  By considering many possible models, Bayesian methods can better reason about uncertainty away from the training data by measuring how highly probably models differ in their predictions.}.

Formulating predictive probability in this way also enables reasoning about uncertainty more in more specific ways than is typically possible with standard probabilistic models.
With a posterior probability, one can compare how different models result in different predictive probabilities.
As an example, in classification if one can \emph{sample} from $p(\theta|\mathcal{D})$ to obtain $n$ different parameter settings $\theta_1, ..., \theta_n$, then they can be used in $n$ different models to produce predictive probabilities $p(y|x,\theta_1), ..., p(y|x,\theta_n)$.
Intuitively, if the predictive probabilities are considerably different across sampled models, then the overall model may be uncertain in the estimate of predictive probability itself.
Here, because the final predictive probability is uncertain due to different sampled models disagreeing in their outputs, this kind of uncertainty is often called \emph{model uncertainty}. 
Model uncertainty metrics such as the variance over sampled predictive probabilities can be used in monitoring to ensure unreliable predictive probability estimates are not used in monitoring models for their reliability during deployment.
\footnote{As a more illustrative example, consider the following case: A system is monitoring the maximum predictive probability from a classifier that uses a posterior in computing $p(y|x)$ as in \eqref{eq:posteriorpp}. If the maximum predictive probability drops, this could indicate that predictions have suddenly become unreliable because the model is signaling that it is uncertain in its predictions.  For this to be useful, the system requires $p(y|x)$ itself to be reliable in order to make such a detection.  If samples from $p(\theta|\mathcal{D})$ result in drastically different $p(y|x,\theta)$ values (i.e. ones with high variance), this can indicate that $p(y|x)$ is unreliable, and the maximum probability is not trustworthy.}. 

It is worth mentioning that in practice Bayesian models come with drawbacks.
For many modern classes of ML models, such as deep neural networks, computing \eqref{eq:posteriorpp} or even the posterior exactly is intractable.
A significant line of research has been dedicated to finding tractable ways of approximating key terms in Bayesian models, including methods for neural networks~\cite{neal1992belief, graves2011variational, blundell2015weight, gal2016dropout, maddox2019simple, daxberger2021laplace, wright2024analytic}.
Approximations often require specialized training and/or inference procedures that can add complexity and computational burden when compared to standard methods.
Finally, while Bayesian neural networks have shown to produce more useful measures of uncertainty, they often report slightly worse predictive accuracy than standard neural networks.
Careful evaluation should be performed to determine if the benefits of using Bayesian models outweighs their detriments.


\textbf{Epistemic and Aleatoric Uncertainty}
Model uncertainty metrics can provide the basis for monitoring models during deployment.
However, in principle $p(y|x)$ can express uncertainty for any number of reasons, not just because an instance is out-of-distribution.
Fortunately, Bayesian models are often equipped with methods to disambiguate sources of uncertainty.
Specifically, many models that are able to express model uncertainty have explicit means to separate \emph{epistemic} and \emph{aleatoric} uncertainty~\cite{der2009aleatory, kendall2017uncertainties, hullermeier2021aleatoric}.
Epistemic uncertainty is often used synonymously in ML with model uncertainty, or uncertainty associated with the estimation of model parameters.
Aleatoric uncertainty is uncertainty caused by inherent noise in the data.

The important practical distinction between the two is that epistemic uncertainty can be reduced by collecting more data while aleatoric uncertainty cannot.
When a model distinguishes between the two, each can be used to determine appropriate courses of action once a model is found to be uncertain.
If a model is found to be epistemically uncertain in a particular failure mode, this indicates that the model may be encountering out-of-distribution data, and developers can collect more data and retrain a model to reduce uncertainty.
If a model is found to be aleatorily uncertain, these are likely instances that are being affecting by some form of noise that makes the relationship between instance and label seem random.
In these cases, there is likely no course of action that can reduce uncertainty, and the broader system in which the model is embedded should handle them
\footnote{In practice, there are a number of different ways to handle noisy data.  Developers often build mechanisms for detecting noise into data pipelines so that the model is not responsible for detecting known noise cases.  Even in cases where the larger system does not detect that an instance is noisy and the model has high aleatoric uncertainty, these instances can be assumed to be failures and risk mitigation strategies (e.g. differing to human judgement) can be employed.}.
Both aleatoric and epistemic uncertainty can be monitored during deployment, and proper courses of action can be built in to avoid failure caused by these sources.
Further, both sources of uncertainty can be used to provide developers information that can help them iterate in model development.

\examplebox{Running Example: Autonomous Vehicle with Visual Obstacle Avoidance}{
    The obstacle avoidance model team finds that monitoring metrics based on predictive probability sometimes fails to flag when their object detector will make an incorrect prediction.
    To remedy this, they train a Bayesian Neural Network version on their model based on ~\cite{kendall2017uncertainties}.
    Instead of the model just expressing predictive probability, it also expresses quantifiable measures of both epistemic and aleatoric uncertainty.
    They choose to monitor these metrics during deployment instead of just predictive probabilities.
    They find that in some cases where the previous detector's predictive probability metrics fail to flag incorrect predictions, the epistemic uncertainty metric correctly detects when the model fails.
    Equipped with a new metric that can monitor for failures, the autonomous vehicle can better utilize built-in redundancies in the system to avoid potential collision with obstacles.

    Upon further investigation, they discover that the model has consistently high epistemic uncertainty when trying to detect garbage trucks.
    Even further investigation reveals that the training set for the model contains no garbage trucks, as their data collection efforts always fell on days when the city was not collecting garbage.
    As a result, the appearance of garbage trucks during deployment represented a shift from the training data.
    They make plans to begin collecting training data on days when city garbage trucks are on the roads in order to remedy the difference in data from training to deployment.
}

\subsection{Perturbation Robustness}%
\label{sec:perturbation_model}
While the Bayesian framework provides general formalisms for reasoning over many possible models to detect out-of-distribution data, it does not provide formalisms that characterize the nature of a shift.
If developers can anticipate what kinds of shifts are likely to occur, they can build these assumptions into how they train and use their models.
One example is based on the observation that in many domains, small changes to instances should not drastically change the true label a model should predict when given the instance as input.
Consider the task of classifying the species of birds in images.
If a relative few pixels of the images are changed, or if all pixels are only slightly changed (e.g. all pixels are slightly more red), the bird in the image will likely not appear more like a different species.
This is the basic assumption behind \emph{perturbation robustness}: Small changes, or perturbations, to an instance should not change the true label of that instance.

The most common formal definition of a perturbation is that given an instance $x$, a perturbation takes the form of $x' = x + \delta$; that is, an instance is perturbed by adding \emph{noise} $\delta$.
Obviously, in order for this definition to be practically useful, some restriction must be placed on $\delta$ otherwise the assumption that the label does not change after the perturbation will not hold
\footnote{For instance, if $\delta = -x$, then there is no way for a model to correctly predict the label for $x'$.}.
For this reason, different formulations of perturbation robustness consider different \emph{classes of perturbations} $\Delta$\footnote{Sometimes called a \emph{perturbation budget.}}, which can restrict $\delta$ based on properties such as its magnitude~\cite{goodfellow2015,gowal2020uncovering,croce2021mind}, the number of elements (such as pixels) in $x$ that it can be applied to~\cite{sharif2016accessorize, papernot2016limitations}, and even if the perturbation is realistic to occur in the physical deployment environment~\cite{eykholt2018robust, tu2020physically}.
Prior research has demonstrated that simple classes of perturbations---even perturbations so small as to be imperceptible to humans---can cause a drastic increase in the failure rate of modern supervised ML models~\cite{szegedy2014,carlini2017, ilyas_adversarial_2019, zhang_theoretically_2019, zou2023universal}.

Fortunately by constraining the kinds of perturbations that can occur to those within a class, we can define robustness in terms of the same formalisms used to define generalization in-distribution, and use these to reason about the robustness properties of models or even create models that are robust to perturbations.
Formally, a model that is robust to a class of perturbations is one that minimizes the following definition of expected error (or risk):
\begin{equation}\label{eq:adversarial_risk}
    \mathbb{E}\left[ \max_{\delta \in \Delta} \ell \left(y, f(x + \delta) \right) \right]
\end{equation}
Here, expected error is defined similarly to \eqref{eq:risk}, but loss is measured over the \emph{worst-case} perturbation in the class of perturbations, i.e. the perturbation that makes the ML model incur the most error.

In defining error in terms of a worst-case perturbation, a model does not explicitly reduce error over all possible perturbations in $\Delta$, just the ones in which the model has the highest error.
This gives rise to the interpretation of \eqref{eq:risk} as a measure of \emph{adversarial error}, as the worst-case can be seen as being chosen by an adversary whose goal is to find a perturbation where the ML model performs the worst.
In practice, this gives rise to methods that need not focus on any perturbation, only adversarial chosen ones.
Also, the adversarial setting allows developers to define the conditions in which a model should be robust to real-world influence by adversarial actors.
Indeed, researchers have shown adversarial attacks derived from this basic framework can result in real-world perturbations that can affect ML-based facial recognition models~\cite{sharif2016accessorize}, road sign detection models in autonomous vehicles~\cite{eykholt2018robust}, and models that classify the sentiment of text~\cite{morris2020textattack}.

Methods to learn or reason about the robustness properties of models to perturbations typically fall into two categories.
First, there are a number of \emph{robust training} techniques that train models with certain classes of perturbations in mind.
These techniques range from new objective functions that take into account the class of perturbations ~\cite{sinha2018certifying} to data augmentation schemes ~\cite{zhao2020maximum} to alternative optimization problems that aim to reduce the ``spurious'' relationships between instances and labels that an adversary can exploit ~\cite{arjovsky2019invariant}.
Second there is an emerging trend to applying \emph{formal verification} to ML models to determine whether they are robust to classes of perturbations ~\cite{katz-etal:CAV:2017:reluplex,DBLP:conf/cav/KatzHIJLLSTWZDK19,bak2021nfm,DBLP:conf/cav/LopezCTJ23,DBLP:journals/corr/abs-2307-10266,DBLP:journals/pacmse/Duong0ND24,zhang2018efficient,xu2020automatic,salman2019convex,xu2021fast,wang2021beta,zhang22babattack,zhang2022general,shi2024genbab,kotha2023provably,DBLP:journals/pacmpl/SinghGPV19}.
Formal verification methods can provide proofs that assure models are robust to classes of perturbations or examples where they are not, and in so providing a powerful tool for evaluating models for their robustness.
In practice, applying these methods have challenges.
First, specifying classes of perturbations in the formalisms required by formal verifiers is non-trivial, and often requires considerable knowledge of both the application domain of the ML model and how formal verifiers work.
Second, many modern ML models are complex enough to pose scalability issues with verifiers, often making formal verification intractable.
Nevertheless, if a formal specification can be written and a model is simple enough so that verifiers can scale, formal verification of robustness is possible.

\subsection{Divergence Robustness}
In contrast to perturbation robustness that defines shifts in distributions through a particular functional form, \emph{divergence robustness} uses probabilistic definitions of how a $\tilde{P}(x,y)$ changes from $P(x,y)$.
To get to useful probabilistic definitions of divergence from training to deployment distributions, we again look to rules of conditional probability to decompose the data generative process into components that can more specifically define what assumptions are made about the nature of a shift.
Three main probabilistic assumptions that are often used to formalize a distributional shift:
\begin{enumerate}
    \item \textbf{Covariate Shift} - The distribution over \emph{instances} changes, but the relationship between instances and labels remains the same. Formally:
    \begin{itemize}
        \item $P(x,y) = p(y|x)p(x)$
        \item $\tilde{P}(x,y) = p(y|x)\tilde{p}(x)$
        \item $p(x) \neq \tilde{p}(x)$
    \end{itemize}
    \item \textbf{Concept Shift} - The distribution over instances stays the same, but the relationship between instances and labels changes. Formally:
    \begin{itemize}
        \item $P(x,y) = p(y|x)p(x)$
        \item $\tilde{P}(x,y) = \tilde{p}(y|x)p(x)$
        \item $p(y|x) \neq \tilde{p}(y|x)$
    \end{itemize}
    \item \textbf{Label Shift} - The distribution over \emph{labels} changes, but the relationship between labels and instances remains the same. Formally:
    \begin{itemize}
        \item $P(x,y) = p(y|x)p(x)$
        \item $\tilde{P}(x,y) = p(x|y)\tilde{p}(y)$
        \item $p(y) \neq \tilde{p}(y)$
    \end{itemize}
\end{enumerate}

Covariate shift (also known as \emph{sample selection bias}) can be practically understood as cases where the nature of how instances are generated by an environment changes.
Rare instances in the training set may be common occurrences in an environment and vice versa.
Classic methods ~\cite{shimodaira2000improving, zadrozny2004learning, sugiyama2007direct, gretton2008covariate} adapt supervised models to covariate shift by \emph{re-weighing} instance/label pairs during training, that is, changing the loss function during training so that some training data are considered ``more important'' during training than others.
Intuitively, these methods give more weight to training instances that have higher $\tilde{p}(x)$, as they are more reflective of data common in the deployment environment.

While classic methods are often simple, effective, and principled, they have practical issues when being applied to many modern supervised ML problems.
First, many of these methods either implicitly or explicitly need to learn a model of $\tilde{p}(x)$, which requires instances from a deployment environment.
In practice, development teams may not know a priori that their deployment environment is generating shifted covariates and would not collect data to model the shift.
Second, in many modern ML applications, instances are \emph{high-dimensional}, which means that an instance contains many values \footnote{More formally, instances are often represented by real-valued vectors, matrices, or tensors.  High-dimensionality means that instances of these forms have many elements.  As an example, an image is often represented as a 3-tensor where the length and width is determined by the resolution of the image.}.
Estimating $\tilde{p}(x)$ (often called a \emph{density}) or even statistics of this distribution for high dimensional $x$ is known to be challenging and can result in poor reweighing ~\cite{sugiyama2012density, stojanov2019low}.

In response to these challenges, recent methods aim to learn $\tilde{p}(x)$ \emph{online} ~\cite{zhang2024adapting}, or as instances are observed and used by a model to make predictions while it is deployed.
This way, a covariate shift can be monitored for and detected in the natural course of model deployment.
If instances are flagged as resulting from a covariate shift, proper caution can be taken when relying on the predictions of a model\footnote{It is important to note that if an instance is known to be out-of-distribution, it does not \emph{necessarily} mean that a model will fail.  It does mean that the instance is unlike the data used to train a model and caution may be warranted, as there might not be a principled way to reason whether the model will fail.  Contrast this to monitoring for reliability.  A calibrated model can quantify the probability of predictions, which much more clearly quantifies the chance of failure.  This highlights the inherent difficulty in handling robustness versus reliability in ML models.}.
Similarly, recent techniques have developed methods for \emph{dimensionality reduction} for the purpose of reweighing training data when instances are high-dimensional ~\cite{stojanov2019low}.

Label shift assumes a data generation process where the label of an instance is generated, then an instance is generated given that distribution.
Here, the distribution that models the probability of a label $p(y)$ changes from train time to deployment.
Rare classes during training can be more common during deployment, or vice versa.
High regression labels that were common during training can be rare during deployment, or vice versa.

A key task in detecting and adjusting models for label shift is estimating $\tilde{p}(y)$.
Some of the earliest work on label shift showed that if this probability is known it can be used to weigh predictions of models to adjust for label shift~\cite{elkan2001foundations} or monitored to detect label shift during deployment~\cite{saerens2002adjusting}.
If samples from $\tilde{p}(y)$ are given, then its estimation is often trivial \footnote{For example, in classification one can simply find the probability of a class by counting the number of times a class label is sampled from $\tilde{p}(y)$, and divide it by the number of total samples.}, but this would require effort to label instances from a deployment environment after a shift has occurred, which is not helpful if trying to detect and mitigate the effects of a possibly unknown shift.
Most methods assume, instead, that they can learn the shift in labels from instances sampled from the $\tilde{P}(x,y)$, which can be obtained from an environment as the model is deployed and do not require labeling.
Early work focused on learning $p(x|y)$ as a means to estimate $\tilde{p}(y)$~\cite{chan2005word, storkey2008training} \footnote{While these and other methods use $p(x|y)$ differently, there is a common intuition as to why learning $p(x|y)$ helps in learning $\tilde{p}(y)$. By the definition of label shift, both the training distribution $P(x,y)$ and the deployment distribution $\tilde{P}(x,y)$ share the same $p(x|y)$.  If $p(x|y)$ is known, then any difference between the two can be attributed to differences in $p(y)$ and $\tilde{p}(y)$.  Thus, estimating $p(x|y)$ can lead to ways of reasoning about shifts in label distributions.}.
However, $p(x|y)$ is a density similar to $p(x)$, and learning it has similar challenges as methods to handle covariate shift.
More recently, the authors of~\cite{lipton2018detecting} developed a principled method for detecting and correcting label shift in models that does not scale with the dimensionality of instances, and thus can be used for high-dimensional learning problems.
Separately, in~\cite{bai2022adapting} the authors propose an online method for adapting to label shift.

Finally, in concept shift (or concept \emph{drift})~\cite{gama2014survey, suarez2023survey} the relationship between labels and instances, as modeled by $p(y|x)$ changes from train time to deployment.
This represents a fundamentally more difficult problem than label or covariate shift as sampling instances or labels from a deployment environment alone do not provide enough information to detect or mitigate the effects of concept drift without strong assumptions.
Under concept shift, an instance can take one class during training and a different one during deployment, leading to a classifier that was trained to make the wrong prediction.

If a monitor is able to receive labels from an environment as a model makes predictions, detecting and adapting to a concept drift has established foundations~\cite{bartlett1992learning} and practical algorithms have been developed~\cite{gama2004learning, harel2014concept, pesaranghader2016fast, hinder2020towards}.
While assuming access to labeled instances during deployment is unrealistic in some settings, there are a number of \emph{data streaming} applications of ML for which this assumption does hold\footnote{For example, models that predict future market trends, or patient outcomes can eventually get correct labels once time catches up to the predictions.}.
Further, many methods benefit from assuming some temporal behavior of the shift, such as a rate, magnitude, or periodicity in which the environment shifts from the training distribution.
With strong assumptions on the behavior of the shift, some methods are able to detect shifts with very few or even no labels in specific learning settings~\cite{awasthi2023theory}.


\subsection{Heuristic Approaches}
In principle, if no strong probabilistic assumptions, behavioral assumptions, or labeled data can be used to characterize a shift from training to deployment distributions, there is no reliable way to generalize or even detect to a shift.
Despite this, there have been a number of empirical results that show some ML models have surprising robustness despite making no explicit assumptions about the nature of shifts they intend to model.
The authors of ~\cite{miller2021accuracy} show that for across many modern neural network models, in-distribution accuracy strongly correlates with out-of-distribution accuracy in a broad set of benchmark supervised learning tasks.
A number of follow-on works studied this observation in different settings ~\cite{ming2022impact, kumar2022fine, baek2022agreement, lee2023surgical, xie2024importance}, which have motivated use of certain techniques that can achieve high in-distribution accuracy but also perform well empirically out-of-distribution.

These findings are seemingly surprising.
Why would a model trained to generalize to one distribution, generalize to an entirely different one?
One way of understanding these results is that some methods for training supervised models contain no explicit modeling of robustness, but perhaps make key \emph{implicit} assumptions that bias models in a way that allows them to generalize to realistic shifts in data distributions.
Such biases can be introduced by design decisions such as the data sets models are trained on, the methods used to train models, and the classes of models used during training.
Even though some these decisions were made to achieve high in-distribution performance, they also result in models that are robust to the kinds of shifts that are the focus on empirical research.

More broadly, these results motivate the efficacy of \emph{heuristic approaches} to robustness that make few or no formal assumptions about the shifts they are robust to.
These are practically useful as they need little or no deployment data to use and typically require no formal analysis of the kinds of shifts the model could encounter during deployment. 
Beyond the techniques mentioned above to train models to be more robust out-of-distribution, a number of heuristics have emerged to detect when an instance is out of distribution.
Simple heuristics such as setting a threshold on the maximum confidence value of a classifier have been shown to be strong baselines for detecting out-of-distribution instances~\cite{hendrycks2017baseline}.
Since then, many heuristics have been developed that utilize properties of models trained in-distribution to perform out-of-distribution detection~\cite{liang2018enhancing, lee2018simple, hsu2020generalized, liu2020energy, sun2021react, sun2022out}.

The downside of heuristic approaches is that their lack of formal modeling makes it difficult to reason about the limits of their robustness properties.
Many of the successes reported by these and related methods are on benchmark data sets where one data set is considered "in-distribution" and a separate data set is considered "out-of-distribution".
Without formal assumptions about the shift being considered, it is unclear how exactly to interpret these results relative to different applications.
For instance, a popular suite of data shift benchmark data sets is WILDS~\cite{koh2021wilds}.
WILDS contains tasks such as image classification of animals from trail cameras where the shift in distribution is represented by the training set containing images from cameras in different geographical locations than the images in the test set. 
If an out-of-distribution detection method performs well on this benchmark, and makes no formal assumptions about the nature of the shift, there is no strong principled reason to believe the model will be as successful on a different shift.
Thus, while many of these techniques have shown impressive empirical success on benchmarks, it's often not clear to development teams if they are appropriate for the tasks they are building ML models for.

\examplebox{Running Example: Autonomous Vehicle with Visual Obstacle Avoidance}{
    The obstacle avoidance model team is concerned that their object detection model is not robust to changes in the kinds of objects their model could encounter during deployment.
    Namely, they find that the training data they collected contains many cars relative to pedestrians or bicycles, as the local city's infrastructure is mainly designed for automobile travel.
    They fear that if they deploy their autonomous vehicles in different cities that have more commuters that walk or cycle, their model may fail.
    They recognize this as a potential label shift they wish to detect.
    For this, they implement a monitor in the autonomous vehicle, based on~\cite{lipton2018detecting}, that performs a statistical test to determine whether a significant shift in labels has occurred.
    When the monitor raises a flag for a significant shift, it enters a manual mode that slows the car down and requires the human passenger to take over control of the vehicle.

    Later, the team finds that someone, who apparently feels negatively about autonomous vehicles, is putting specifically designed stickers on lamp posts that make their model sometimes fail to detect them.
    If their model fails to detect lamp posts when the vehicle has to make tight turns, it can get dangerously close to colliding with them.
    To mitigate the effects of this adversarial attack, the team employs a training technique that can provide certified robustness against such stickers~\cite{chiang2020certified}.
    Not only do they find that their model is more robust to the stickers placed on lamp posts, but it is also more robust to graffiti and other markings.

    Because they recognize that other, unexpected shifts in the deployment distribution are possible, the team wants to employ heuristic approaches to detect other shifts that may not be caught by other monitors.
    For this they implement a version of a heuristic approach~\cite{sun2022out}, to determine if new instances are much different from a sampling of the instances used during training.
    If new samples are significantly different from training examples, the car enters manual mode until instances it receives are sufficiently similar to training instances.

    It is worth noting that a significant amount of efficiency engineering, testing, tuning, and system design is required to make these monitors work in practice.
    For example, to utilize the heuristic posed in~\cite{sun2022out}, distance must be computed between new instances being observed by the model during deployment and instances used in training.
    Performing so many distance computations may not be possible in real-time on the kind of hardware that can be put in an autonomous vehicle.
    Many design decisions and optimizations may be necessary, such as efficient algorithms for computing pair-wise distances~\cite{johnson2019billion}, and ways of sub-sampling training data so all training instances need not be in the vehicle's memory. 
    Further, it is not obvious how dissimilar an instance observed during deployment must be to cause a likely failure.
    Careful test and evaluation should be done to determine proper thresholds to be used during monitoring.
    Finally, how the autonomous vehicle should react when the monitor determines that an instance is out-of-distribution should be carefully considered in relation to the risks associated with the model failing.
    Proper system design and testing should be performed to achieve an acceptable level of vehicle autonomy and safety.
    }

\subsection{Engineering Implications}
Robustness poses unique challenges for developers both before and after deployment of ML models.
Because there are many ways data can shift from the training distribution, developers must make decisions about what kinds of shifts they will design their ML-enabled systems to be robust for, and employ the techniques to mitigate them.
Prior sections discussed some techniques to train models that can be robust against shifts (e.g. robust training against adversarial attacks or label shifts), as well as metrics that can be used to monitor models for a shift during deployment (e.g. epistemic uncertainty or heuristic out-of-distribution detection).
The choice of techniques used for both training and monitoring are dependent on what developers assume to be important robustness cases in their application.
If it assumed that adversaries may use ``patch'' attacks against a computer vision model, techniques for training models robust against adversarial patches should be employed~\cite{xiang2021patchguard, chen2022towards}.
If there is a concern that instances will change over time in an environment, then methods for monitoring for covariate shift should be employed~\cite{jang2022sequential, zhang2024adapting}.
If there is no strong sense of the kinds of shifts that can happen heuristic approaches can be used to monitor for shifts.
Ultimately, it is up to developers to determine what kinds of shifts are realistic and utilize the proper techniques for robust training or monitoring for shifts.

Testing models for their robustness presents a related set of challenges.
The case-based testing framework outlined in Sec. \ref{sec:empiricallyevaluating} requires labeled examples to evaluate model performance.
ML model robustness concerns itself with deployment scenarios where no labeled data is available to even train from.
This ``catch-22'' makes true robustness testing of ML models seem impossible: How does one test for scenarios that have no test data?
In practice, developers can make assumptions about the nature of shifts that can occur and build test cases around them.
Developers can intentionally leave labeled data that represent important scenarios out of the training set of a model and build test cases around those.
If the model passes these cases, this can provide evidence that a model generalizes to important scenarios outside its training set, and possibly others as well
\footnote{It is important to re-emphasize that, in principle, if a model does indeed generalize to out-of-distribution data, then some implicit or explicit assumption of the model has allowed it to do so.  Without some understanding of why this occurred, it's unclear how exactly to interpret the results of such robustness tests outside the specific cases they were designed for.}.
Even if data from new deployment scenarios cannot be gathered for tests, data augmentations can be applied to data to create such scenarios.
For example, developers can inject noise during testing to evaluate a model's robustness in simulated conditions or environments.
Much like when choosing which robust training or monitoring techniques to use, developers must consider the possible robustness scenarios a model may encounter and build test cases around them.

\examplebox{Running Example: Autonomous Vehicle with Visual Obstacle Avoidance}{
    A team working on sensors for the autonomous vehicle notifies the obstacle avoidance model team that the visual sensor on the vehicle can sometimes drop a color channel~\footnote{Visual imagery is often represented by three ``channels'' for the red, blue, and green values for each pixel.  If one is ``dropped'', then the image contains no values for that color.} when sending data to the obstacle avoidance model.
    The obstacle avoidance model team recognizes this as possible out-of-distribution data that their model could encounter during deployment.
    They consider retraining their model on augmented data with channels dropped in order to be robust to this possible scenario.
    Before they do, they choose to create a test case consisting of images with channels randomly dropped through augmentation.
    They find that their model passes this test case above the acceptance threshold defined on nominal data.
    Because of this, they determine that retraining is not necessary and include the ``channel-drop augmentation'' test case in their suite of tests going forward.
    }

\subsection{Summary}
In this section, we defined robustness of a model to avoid failure when it receives out-of-distribution data.
In principle, no model can be robust to a shift in distribution without explicit or implicit assumptions that guide how a model can generalize to out-of-distribution data.
Because of this, there are a number of formal models of the kinds of shifts that can occur, and techniques to either detect when a shift will occur or to adapt a model to shifts.
In practice, developers must understand the application domain of the models they build to define what kinds of robustness they need to consider.
Then, they can implement techniques to train robust models, monitor for shifts in distribution, and create test cases that evaluate models for their robustness in critical scenarios.

\section{Conclusion}
\label{sec:conclusions}
Though the number of applications in which machine learning represents the state-of-the-art is growing swiftly, ML models are among the most difficult software to reason about their potential for failure.
There exists numerous possible causes of ML model failure, ranging from those stemming from principled assumptions being violated to implementation errors.
In this guide, we reviewed a large body of work that considers these reasons for failure and surveyed techniques that can be used to reduce or mitigate the effects of ML model failure.
We structured this guide around the concepts of reliability and robustness, as the distinction between the two highlights an important formal principle that explains how ML models are able to generalize, and thus motivate different ways of reasoning about failure.
Subjects covered in this guide represent a broad body of ML research, but were tied together in practical activities that developers should consider when building ML models.
Specifically, we discussed how principles behind ML model reliability and robustness motivate practices and techniques related to data set collection, test and evaluation, training, and monitoring of ML models.
This guide is by no means comprehensive, and many of the running examples used are overly simplistic.
To gain a more complete, current understanding of how ML models can fail and the techniques used to mitigate failures, a number of other topics should be understood.
Below we discuss a few additional considerations not considered in this guide.

\textbf{The Practice of Building Machine Learning Models} When building ML models for real-world applications there are often sophisticated pipelines that take developers from initial concept to a final model embedded in a system.
The number of tasks within this pipeline are many, and few were discussed in detail in this guide.
Specifically, we did not discuss the significant amount of trade-craft and numerous best practices that go into training ML models.
Teams must quickly iterate on training, evaluating, and adjusting their procedures before even producing the first model worthy of test and evaluation.
Depending on the application, teams may require knowledge of specialized hardware, numerical optimization, and deep understanding of the models themselves to produce a viable model.
Some of this is covered in formal education in ML or in various guides~\cite{bengio2012practical,parthasarathy2024ultimate}, but much of it is scattered across many sources.
The reason why this is important in the context of this guide is that \emph{any error in the practice of building ML model may cause failure even if the principles are well understood}.
As such, beginning from strong ML practices is key to ensuring models do not fail unexpectedly.

\textbf{System-level Reliability and Robustness}
One of the reoccurring themes in this guide is that ML model failure should be understood in context.
That is, ML model outputs should be considered with respect to the expectations of the model.
More broadly, \emph{failure should be considered not just at the level of an ML model, but the system in which an ML model is embedded}.
The running example of an autonomous vehicle is illustrative of this point.
An ML model may fail to detect a pedestrian, but the vehicle may still safely navigate around them.
Conversely, the model may detect a pedestrian but not do so in a timely enough fashion for the vehicle to avoid them.
Focusing on the reliability and robustness of the systems and components around ML models is critical to practically understanding and mitigating failures.
Fortunately, some technical communities are taking a systems-level view of reliability and robustness of ML-enabled systems.
For instance, the robotics community has a great deal of work of how ML models influence the robustness of overall robotic systems ~\cite{pezzementi2018putting,hawkins2021guidance,abrecht2021testing,weyns2023towards}.

\textbf{Operational Design Domains for ML-Enabled Systems}
For many engineers, our definitions of reliability and robustness may not be satisfying.
Defining a specification as being a probability distribution can lead to an unacceptable amount of ambiguity.
While these definitions allow us to tie formal principles of ML to engineering concepts, there may be better ways in practice to define what is "in-specification" for ML models.
Parts of the autonomous vehicle community have adopted the concept of an Operational Design Domain (ODD)~\cite{czarnecki2018operational, koopman2019many} as a way to define the conditions in which autonomous functionality in vehicles can be safely turned on.
ODDs describe specifications in terms of environmental factors, which can place failure more solidly in the environment in which systems are deployed, thus enabling better understanding of their practical limitations.
A prevailing challenge faced by the ML and AI research communities is how to tie the theory of ML failure to more practical specifications of environments such as those provided in an ODD.
If such theory is established, more rigorous and principled practices can be developed to better understand failure in ML-enabled systems.

\textbf{Human/Computer Interaction}
Even if an ML model is successfully integrated into a larger system, in many cases the system is used by a human user.
If user is unsuccessful at the task in which the ML-enabled system is tasked to help them with, it can sometimes be traced back to some fundamental mismatch between the system's assumptions and the user's expectations.
This is why the interactions between an ML-system and its end-users must also be considered when reasoning about failures.
Human/Computer interaction (HCI) is a large technical field with a considerable amount of research and numerous practical applications, but ML-enabled systems can pose unique challenges to users not present in other systems.
This has lead to specialized ML topics such as \emph{interpretability} and \emph{explainability}~\cite{gilpin2018explaining, burkart2021survey}, which aim at providing humans better understanding of ML models
\footnote{Maintaining consistent definitions has proved challenging in research communities that focus on interpretability and explainability.  Some works use them interchangeably, while others draw important distinctions between them.  For instance, the authors of~\cite{gilpin2018explaining} define interpretability as ``the science of comprehending what a model did (or might have done)'' and explainability as ``(the ability for a model) to summarize the reasons for (model) behavior, gain the trust of users, or produce insights about the causes of their decisions.''  While~\cite{rudin2019stop} differentiates the two as interpretability being an inherent property of the model, while explainability is something external to the model.  As such, it is difficult to provide a more specific definition than the one we provide here: The goal of explainabilty/interpretability is to provide humans some form of understanding of an ML model}.
In fact, some work has focused on how humans can reason about failure using some of the formalisms discussed in this guide~\cite{prabhudesai2023understanding}.
\emph{Understanding the specific goals and decision-making processes that users undergo with aid of the ML-enabled system, as well as how the system interacts with the user is important in practically understanding failures in how ML models are used}
\footnote{It is worth noting that interpretability and explainability have become somewhat controversial topics.  Some of the basic assumptions made by important techniques have been called into question~\cite{lipton2018detecting,rudin2019stop}.  A full treatment of explanability and interpretability is out of the scope of this guide.  We simply wish to emphasize the need for human interaction to be considered when reasoning about possible failures in ML model deployment.}.

\textbf{Foundation and Modern Generative Models}
In this guide, we considered the traditional supervised learning setting where models are trained using only a fixed training set that defines what is ``in-distribution.''
The distinction between data that is in-distribution versus what is out-of-distribution becomes more complex when considering some more recently adopted ways that machine learning models are built.
For instance, many computer vision and language models are not trained solely from a fixed training set, but instead adapt a pre-trained model called a \emph{foundation model} through a process called \emph{fine-tuning}.
The foundation model was created using data separate from what is used to fine-tune it.
As such, the data set used to fine-tune a model is not the only data that influences the final choice of model made by the training algorithm.
This is further complicated when considering common ways in which modern generative models are trained.
Most commonly used Large Language Models such as Llama~\cite{touvron2023llama,touvron2023llama2,dubey2024llama} are pretrained using massive, web-scale data sets and further adapted using multiple different training approaches before developers can fine-tune the models for their specific purposes.
As such, it is much less clear what data is actually out-of-distribution.
The practical robustness benefits of using pre-trained foundation models as a basis for fine-tuning has been shown empirically.
Specifically, results indicate that models fine-tuned from foundation models generalize better to out-of-distribution data than models that were trained using more traditional supervised learning approaches~\cite{oquab2024dinov2,sun2024evaluating}, which indicates that generalization outside of data used to fine-tune a foundation model may be possible through pre-training.
Many of the techniques discussed above still apply to models fine-tuned from foundation models, but it is worth noting that \emph{some of the foundational assumptions behind generalization in supervised learning are fundamentally different in this training paradigm, and new foundations may be required to formally understand the scope of how these models generalize}.

\section*{Acknowledgements}
Copyright 2025 Carnegie Mellon University.

This material is based upon work funded and supported by the Department of Defense under Contract No. FA8702-15-D-0002 with Carnegie Mellon University for the operation of the Software Engineering Institute, a federally funded research and development center.  

The view, opinions, and/or findings contained in this material are those of the author(s) and should not be construed as an official Government position, policy, or decision, unless designated by other documentation.

[DISTRIBUTION STATEMENT A] This material has been approved for public release and unlimited distribution.  Please see Copyright notice for non-US Government use and distribution.

This work is licensed under a Creative Commons Attribution-NonCommercial 4.0 International License.  Requests for permission for non-licensed uses should be directed to the Software Engineering Institute at permission@sei.cmu.edu.

Carnegie Mellon® is registered in the U.S. Patent and Trademark Office by Carnegie Mellon University.

DM25-0260

\bibliographystyle{plain}
\bibliography{references}

\begin{thebibliography}{100}

\bibitem{abrecht2021testing}
Stephanie Abrecht, Lydia Gauerhof, Christoph Gladisch, Konrad Groh, Christian Heinzemann, and Matthias Woehrle.
\newblock Testing deep learning-based visual perception for automated driving.
\newblock {\em ACM Transactions on Cyber-Physical Systems (TCPS)}, 5(4):1--28, 2021.

\bibitem{evidently2024model}
Evidently AI.
\newblock Model monitoring for ml in production: a comprehensive guide.
\newblock \url{https://www.evidentlyai.com/ml-in-production/model-monitoring}, 2024.
\newblock Accessed: 2024-12-08.

\bibitem{aiid:4}
AIAAIC.
\newblock Incident number 4.
\newblock {\em AI Incident Database}, 2018.

\bibitem{aminikhanghahi2017survey}
Samaneh Aminikhanghahi and Diane~J Cook.
\newblock A survey of methods for time series change point detection.
\newblock {\em Knowledge and information systems}, 51(2):339--367, 2017.

\bibitem{angelopoulos2021uncertainty}
Anastasios~Nikolas Angelopoulos, Stephen Bates, Michael Jordan, and Jitendra Malik.
\newblock Uncertainty sets for image classifiers using conformal prediction.
\newblock In {\em International Conference on Learning Representations}, 2021.

\bibitem{arjovsky2019invariant}
Martin Arjovsky, L{\'e}on Bottou, Ishaan Gulrajani, and David Lopez-Paz.
\newblock Invariant risk minimization.
\newblock {\em arXiv preprint arXiv:1907.02893}, 2019.

\bibitem{awasthi2023theory}
Pranjal Awasthi, Corinna Cortes, and Christopher Mohri.
\newblock Theory and algorithm for batch distribution drift problems.
\newblock In {\em International Conference on Artificial Intelligence and Statistics}, pages 9826--9851. PMLR, 2023.

\bibitem{baek2022agreement}
Christina Baek, Yiding Jiang, Aditi Raghunathan, and J~Zico Kolter.
\newblock Agreement-on-the-line: Predicting the performance of neural networks under distribution shift.
\newblock {\em Advances in Neural Information Processing Systems}, 35:19274--19289, 2022.

\bibitem{bai2022adapting}
Yong Bai, Yu-Jie Zhang, Peng Zhao, Masashi Sugiyama, and Zhi-Hua Zhou.
\newblock Adapting to online label shift with provable guarantees.
\newblock {\em Advances in Neural Information Processing Systems}, 35:29960--29974, 2022.

\bibitem{bak2021nfm}
Stanley Bak.
\newblock nnenum: Verification of relu neural networks with optimized abstraction refinement.
\newblock In {\em NASA Formal Methods Symposium}, pages 19--36. Springer, 2021.

\bibitem{bartlett1992learning}
Peter~L Bartlett.
\newblock Learning with a slowly changing distribution.
\newblock In {\em Proceedings of the fifth annual workshop on Computational learning theory}, pages 243--252, 1992.

\bibitem{bartlett2019nearly}
Peter~L Bartlett, Nick Harvey, Christopher Liaw, and Abbas Mehrabian.
\newblock Nearly-tight vc-dimension and pseudodimension bounds for piecewise linear neural networks.
\newblock {\em Journal of Machine Learning Research}, 20(63):1--17, 2019.

\bibitem{bartlett2008classification}
Peter~L Bartlett and Marten~H Wegkamp.
\newblock Classification with a reject option using a hinge loss.
\newblock {\em Journal of Machine Learning Research}, 9(8), 2008.

\bibitem{bengio2012practical}
Yoshua Bengio.
\newblock Practical recommendations for gradient-based training of deep architectures.
\newblock In {\em Neural networks: Tricks of the trade: Second edition}, pages 437--478. Springer, 2012.

\bibitem{bhardwaj2024machine}
Eshta Bhardwaj, Harshit Gujral, Siyi Wu, Ciara Zogheib, Tegan Maharaj, and Christoph Becker.
\newblock Machine learning data practices through a data curation lens: An evaluation framework.
\newblock In {\em The 2024 ACM Conference on Fairness, Accountability, and Transparency}, pages 1055--1067, 2024.

\bibitem{blundell2015weight}
Charles Blundell, Julien Cornebise, Koray Kavukcuoglu, and Daan Wierstra.
\newblock Weight uncertainty in neural network.
\newblock In {\em International conference on machine learning}, pages 1613--1622. PMLR, 2015.

\bibitem{breck2017ml}
Eric Breck, Shanqing Cai, Eric Nielsen, Michael Salib, and D~Sculley.
\newblock The ml test score: A rubric for ml production readiness and technical debt reduction.
\newblock In {\em 2017 IEEE international conference on big data (big data)}, pages 1123--1132. IEEE, 2017.

\bibitem{buolamwini2018gender}
Joy Buolamwini and Timnit Gebru.
\newblock Gender shades: Intersectional accuracy disparities in commercial gender classification.
\newblock In {\em Conference on fairness, accountability and transparency}, pages 77--91. PMLR, 2018.

\bibitem{burkart2021survey}
Nadia Burkart and Marco~F Huber.
\newblock A survey on the explainability of supervised machine learning.
\newblock {\em Journal of Artificial Intelligence Research}, 70:245--317, 2021.

\bibitem{canal2024decision}
Gregory Canal, Vladimir Leung, Philip Sage, Eric Heim, I~Wang, et~al.
\newblock A decision-driven methodology for designing uncertainty-aware ai self-assessment.
\newblock {\em arXiv preprint arXiv:2408.01301}, 2024.

\bibitem{carlini2017}
N.~Carlini and D.~Wagner.
\newblock Towards evaluating the robustness of neural networks.
\newblock In {\em 2017 IEEE Symposium on Security and Privacy (SP)}, pages 39--57, Los Alamitos, CA, USA, may 2017. IEEE Computer Society.

\bibitem{chan2005word}
Yee~Seng Chan and Hwee~Tou Ng.
\newblock Word sense disambiguation with distribution estimation.
\newblock In {\em IJCAI}, volume~5, pages 1010--5, 2005.

\bibitem{chandola2009anomaly}
Varun Chandola, Arindam Banerjee, and Vipin Kumar.
\newblock Anomaly detection: A survey.
\newblock {\em ACM computing surveys (CSUR)}, 41(3):1--58, 2009.

\bibitem{chandrasekaran2023test}
Jaganmohan Chandrasekaran, Tyler Cody, Nicola McCarthy, Erin Lanus, and Laura Freeman.
\newblock Test \& evaluation best practices for machine learning-enabled systems.
\newblock {\em arXiv preprint arXiv:2310.06800}, 2023.

\bibitem{chandrasekaran2024leveraging}
Jaganmohan Chandrasekaran, Erin Lanus, Tyler Cody, Laura~J Freeman, Raghu~N Kacker, MS~Raunak, and D~Richard Kuhn.
\newblock Leveraging combinatorial coverage in the machine learning product lifecycle.
\newblock {\em Computer}, 57(7):16--26, 2024.

\bibitem{chawla2002smote}
Nitesh~V Chawla, Kevin~W Bowyer, Lawrence~O Hall, and W~Philip Kegelmeyer.
\newblock Smote: synthetic minority over-sampling technique.
\newblock {\em Journal of artificial intelligence research}, 16:321--357, 2002.

\bibitem{chen2022towards}
Zhaoyu Chen, Bo~Li, Jianghe Xu, Shuang Wu, Shouhong Ding, and Wenqiang Zhang.
\newblock Towards practical certifiable patch defense with vision transformer.
\newblock In {\em Proceedings of the IEEE/CVF Conference on Computer Vision and Pattern Recognition}, pages 15148--15158, 2022.

\bibitem{chiang2020certified}
Ping-yeh Chiang, Renkun Ni, Ahmed Abdelkader, Chen Zhu, Christoph Studer, and Tom Goldstein.
\newblock Certified defenses for adversarial patches.
\newblock In {\em 8th International Conference on Learning Representations (ICLR 2020)(virtual)}. International Conference on Learning Representations, 2020.

\bibitem{cortes2016learning}
Corinna Cortes, Giulia DeSalvo, and Mehryar Mohri.
\newblock Learning with rejection.
\newblock In {\em Algorithmic Learning Theory: 27th International Conference, ALT 2016, Bari, Italy, October 19-21, 2016, Proceedings 27}, pages 67--82. Springer, 2016.

\bibitem{croce2021mind}
Francesco Croce and Matthias Hein.
\newblock Mind the box: $ l\_1 $-apgd for sparse adversarial attacks on image classifiers.
\newblock In {\em International Conference on Machine Learning}, pages 2201--2211. PMLR, 2021.

\bibitem{czarnecki2018operational}
Krzysztof Czarnecki.
\newblock Operational design domain for automated driving systems.
\newblock {\em Taxonomy of Basic Terms “, Waterloo Intelligent Systems Engineering (WISE) Lab, University of Waterloo, Canada}, 1, 2018.

\bibitem{dalrymple2024towards}
David Dalrymple, Joar Skalse, Yoshua Bengio, Stuart Russell, Max Tegmark, Sanjit Seshia, Steve Omohundro, Christian Szegedy, Ben Goldhaber, Nora Ammann, et~al.
\newblock Towards guaranteed safe ai: A framework for ensuring robust and reliable ai systems.
\newblock {\em arXiv preprint arXiv:2405.06624}, 2024.

\bibitem{david2010impossibility}
Shai~Ben David, Tyler Lu, Teresa Luu, and D{\'a}vid P{\'a}l.
\newblock Impossibility theorems for domain adaptation.
\newblock In {\em Proceedings of the Thirteenth International Conference on Artificial Intelligence and Statistics}, pages 129--136. JMLR Workshop and Conference Proceedings, 2010.

\bibitem{daxberger2021laplace}
Erik Daxberger, Agustinus Kristiadi, Alexander Immer, Runa Eschenhagen, Matthias Bauer, and Philipp Hennig.
\newblock Laplace redux-effortless bayesian deep learning.
\newblock {\em Advances in Neural Information Processing Systems}, 34:20089--20103, 2021.

\bibitem{der2009aleatory}
Armen Der~Kiureghian and Ove Ditlevsen.
\newblock Aleatory or epistemic? does it matter?
\newblock {\em Structural safety}, 31(2):105--112, 2009.

\bibitem{du2019gradient}
Simon Du, Jason Lee, Haochuan Li, Liwei Wang, and Xiyu Zhai.
\newblock Gradient descent finds global minima of deep neural networks.
\newblock In {\em International conference on machine learning}, pages 1675--1685. PMLR, 2019.

\bibitem{dubey2024llama}
Abhimanyu Dubey, Abhinav Jauhri, Abhinav Pandey, Abhishek Kadian, Ahmad Al-Dahle, Aiesha Letman, Akhil Mathur, Alan Schelten, Amy Yang, Angela Fan, et~al.
\newblock The llama 3 herd of models.
\newblock {\em arXiv preprint arXiv:2407.21783}, 2024.

\bibitem{DBLP:journals/corr/abs-2307-10266}
Hai Duong, Linhan Li, ThanhVu Nguyen, and Matthew~B. Dwyer.
\newblock A {DPLL(T)} framework for verifying deep neural networks.
\newblock {\em CoRR}, abs/2307.10266, 2023.

\bibitem{DBLP:journals/pacmse/Duong0ND24}
Hai Duong, Dong Xu, ThanhVu Nguyen, and Matthew~B. Dwyer.
\newblock Harnessing neuron stability to improve {DNN} verification.
\newblock {\em Proc. {ACM} Softw. Eng.}, 1({FSE}):859--881, 2024.

\bibitem{ebert2016devops}
Christof Ebert, Gorka Gallardo, Josune Hernantes, and Nicolas Serrano.
\newblock Devops.
\newblock {\em IEEE software}, 33(3):94--100, 2016.

\bibitem{eck2022monitoring}
Bradley Eck, Duygu Kabakci-Zorlu, Yan Chen, France Savard, and Xiaowei Bao.
\newblock A monitoring framework for deployed machine learning models with supply chain examples.
\newblock In {\em 2022 IEEE International Conference on Big Data (Big Data)}, pages 2231--2238. IEEE, 2022.

\bibitem{elkan2001foundations}
Charles Elkan.
\newblock The foundations of cost-sensitive learning.
\newblock In {\em International joint conference on artificial intelligence}, volume~17, pages 973--978. Lawrence Erlbaum Associates Ltd, 2001.

\bibitem{eykholt2018robust}
Kevin Eykholt, Ivan Evtimov, Earlence Fernandes, Bo~Li, Amir Rahmati, Chaowei Xiao, Atul Prakash, Tadayoshi Kohno, and Dawn Song.
\newblock Robust physical-world attacks on deep learning visual classification.
\newblock In {\em Proceedings of the IEEE conference on computer vision and pattern recognition}, pages 1625--1634, 2018.

\bibitem{farquhar2021statistical}
Sebastian Farquhar, Yarin Gal, and Tom Rainforth.
\newblock On statistical bias in active learning: How and when to fix it.
\newblock {\em International Conference on Learning Representations}, 2021.

\bibitem{feffer2024red}
Michael Feffer, Anusha Sinha, Wesley~H Deng, Zachary~C Lipton, and Hoda Heidari.
\newblock Red-teaming for generative ai: Silver bullet or security theater?
\newblock In {\em Proceedings of the AAAI/ACM Conference on AI, Ethics, and Society}, volume~7, pages 421--437, 2024.

\bibitem{feiler2006architecture}
Peter~H Feiler, David~P Gluch, and John Hudak.
\newblock The architecture analysis \& design language (aadl): An introduction.
\newblock 2006.

\bibitem{friedenthal2014practical}
Sanford Friedenthal, Alan Moore, and Rick Steiner.
\newblock {\em A practical guide to SysML: the systems modeling language}.
\newblock Morgan Kaufmann, 2014.

\bibitem{gal2016dropout}
Yarin Gal and Zoubin Ghahramani.
\newblock Dropout as a {B}ayesian approximation: Representing model uncertainty in deep learning.
\newblock In {\em Proceedings of the 33rd International Conference on International Conference on Machine Learning - Volume 48}, ICML'16, page 1050–1059. JMLR.org, 2016.

\bibitem{gama2004learning}
Joao Gama, Pedro Medas, Gladys Castillo, and Pedro Rodrigues.
\newblock Learning with drift detection.
\newblock In {\em Advances in Artificial Intelligence--SBIA 2004: 17th Brazilian Symposium on Artificial Intelligence, Sao Luis, Maranhao, Brazil, September 29-Ocotber 1, 2004. Proceedings 17}, pages 286--295. Springer, 2004.

\bibitem{gama2014survey}
Jo{\~a}o Gama, Indr{\.e} {\v{Z}}liobait{\.e}, Albert Bifet, Mykola Pechenizkiy, and Abdelhamid Bouchachia.
\newblock A survey on concept drift adaptation.
\newblock {\em ACM computing surveys (CSUR)}, 46(4):1--37, 2014.

\bibitem{garg2021leveraging}
Saurabh Garg, Sivaraman Balakrishnan, Zachary~Chase Lipton, Behnam Neyshabur, and Hanie Sedghi.
\newblock Leveraging unlabeled data to predict out-of-distribution performance.
\newblock In {\em NeurIPS Workshop on Distribution Shifts: Connecting Methods and Applications}, 2021.

\bibitem{geraci1991ieee}
Anne Geraci.
\newblock {\em IEEE standard computer dictionary: Compilation of IEEE standard computer glossaries}.
\newblock IEEE Press, 1991.

\bibitem{ghosh2022adafocal}
Arindam Ghosh, Thomas Schaaf, and Matthew Gormley.
\newblock Adafocal: Calibration-aware adaptive focal loss.
\newblock {\em Advances in Neural Information Processing Systems}, 35:1583--1595, 2022.

\bibitem{ghosh2023improving}
Subhankar Ghosh, Taha Belkhouja, Yan Yan, and Janardhan~Rao Doppa.
\newblock Improving uncertainty quantification of deep classifiers via neighborhood conformal prediction: Novel algorithm and theoretical analysis.
\newblock In {\em Proceedings of the AAAI Conference on Artificial Intelligence}, volume~37, pages 7722--7730, 2023.

\bibitem{gilpin2018explaining}
Leilani~H Gilpin, David Bau, Ben~Z Yuan, Ayesha Bajwa, Michael Specter, and Lalana Kagal.
\newblock Explaining explanations: An overview of interpretability of machine learning.
\newblock In {\em 2018 IEEE 5th International Conference on data science and advanced analytics (DSAA)}, pages 80--89. IEEE, 2018.

\bibitem{gneiting2007probabilistic}
Tilmann Gneiting, Fadoua Balabdaoui, and Adrian~E Raftery.
\newblock Probabilistic forecasts, calibration and sharpness.
\newblock {\em Journal of the Royal Statistical Society Series B: Statistical Methodology}, 69(2):243--268, 2007.

\bibitem{golowich2018size}
Noah Golowich, Alexander Rakhlin, and Ohad Shamir.
\newblock Size-independent sample complexity of neural networks.
\newblock In {\em Conference On Learning Theory}, pages 297--299. PMLR, 2018.

\bibitem{goodfellow2015}
Ian~J. Goodfellow, Jonathon Shlens, and Christian Szegedy.
\newblock Explaining and harnessing adversarial examples.
\newblock In Yoshua Bengio and Yann LeCun, editors, {\em 3rd International Conference on Learning Representations, {ICLR} 2015, San Diego, CA, USA, May 7-9, 2015, Conference Track Proceedings}, 2015.

\bibitem{gowal2020uncovering}
Sven Gowal, Chongli Qin, Jonathan Uesato, Timothy Mann, and Pushmeet Kohli.
\newblock Uncovering the limits of adversarial training against norm-bounded adversarial examples.
\newblock {\em arXiv preprint arXiv:2010.03593}, 2020.

\bibitem{graves2011variational}
Alex Graves.
\newblock Practical variational inference for neural networks.
\newblock In J.~Shawe-Taylor, R.~Zemel, P.~Bartlett, F.~Pereira, and K.Q. Weinberger, editors, {\em Advances in Neural Information Processing Systems}, volume~24, page 2348–2356. Curran Associates, Inc., 2011.

\bibitem{gretton2008covariate}
Arthur Gretton, Alex Smola, Jiayuan Huang, Marcel Schmittfull, Karsten Borgwardt, and Bernhard Sch{\"o}lkopf.
\newblock Covariate shift by kernel mean matching.
\newblock 2008.

\bibitem{guo2017calibration}
Chuan Guo, Geoff Pleiss, Yu~Sun, and Kilian~Q Weinberger.
\newblock On calibration of modern neural networks.
\newblock In {\em International conference on machine learning}, pages 1321--1330. PMLR, 2017.

\bibitem{gupta2020distribution}
Chirag Gupta, Aleksandr Podkopaev, and Aaditya Ramdas.
\newblock Distribution-free binary classification: prediction sets, confidence intervals and calibration.
\newblock {\em Advances in Neural Information Processing Systems}, 33:3711--3723, 2020.

\bibitem{harel2014concept}
Maayan Harel, Shie Mannor, Ran El-Yaniv, and Koby Crammer.
\newblock Concept drift detection through resampling.
\newblock In {\em International conference on machine learning}, pages 1009--1017. PMLR, 2014.

\bibitem{hawkins2021guidance}
Richard Hawkins, Colin Paterson, Chiara Picardi, Yan Jia, Radu Calinescu, and Ibrahim Habli.
\newblock Guidance on the assurance of machine learning in autonomous systems (amlas).
\newblock {\em arXiv preprint arXiv:2102.01564}, 2021.

\bibitem{heek2018well}
Jonathan Heek.
\newblock Well-calibrated bayesian neural networks.
\newblock {\em University of Cambridge}, 2018.

\bibitem{hendrycks2017baseline}
Dan Hendrycks and Kevin Gimpel.
\newblock A baseline for detecting misclassified and out-of-distribution examples in neural networks.
\newblock In {\em International Conference on Learning Representations}, 2017.

\bibitem{hinder2020towards}
Fabian Hinder, Andr{\'e} Artelt, and Barbara Hammer.
\newblock Towards non-parametric drift detection via dynamic adapting window independence drift detection (dawidd).
\newblock In {\em International Conference on Machine Learning}, pages 4249--4259. PMLR, 2020.

\bibitem{hopkins2023designing}
Aspen Hopkins, Fred Hohman, Luca Zappella, Xavier~Suau Cuadros, and Dominik Moritz.
\newblock Designing data: Proactive data collection and iteration for machine learning.
\newblock {\em arXiv preprint arXiv:2301.10319}, 2023.

\bibitem{hort2024bias}
Max Hort, Zhenpeng Chen, Jie~M Zhang, Mark Harman, and Federica Sarro.
\newblock Bias mitigation for machine learning classifiers: A comprehensive survey.
\newblock {\em ACM Journal on Responsible Computing}, 1(2):1--52, 2024.

\bibitem{hsu2020generalized}
Yen-Chang Hsu, Yilin Shen, Hongxia Jin, and Zsolt Kira.
\newblock Generalized odin: Detecting out-of-distribution image without learning from out-of-distribution data.
\newblock In {\em Proceedings of the IEEE/CVF conference on computer vision and pattern recognition}, pages 10951--10960, 2020.

\bibitem{hullermeier2021aleatoric}
Eyke H{\"u}llermeier and Willem Waegeman.
\newblock Aleatoric and epistemic uncertainty in machine learning: An introduction to concepts and methods.
\newblock {\em Machine learning}, 110(3):457--506, 2021.

\bibitem{hwa2001minimizing}
Rebecca Hwa.
\newblock On minimizing training corpus for parser acquisition.
\newblock In {\em Proceedings of the ACL 2001 Workshop on Computational Natural Language Learning (ConLL)}, 2001.

\bibitem{ilyas_adversarial_2019}
Andrew Ilyas, Shibani Santurkar, Dimitris Tsipras, Logan Engstrom, Brandon Tran, and Aleksander Madry.
\newblock Adversarial {Examples} {Are} {Not} {Bugs}, {They} {Are} {Features}.
\newblock In {\em Advances in {Neural} {Information} {Processing} {Systems}}, volume~32. Curran Associates, Inc., 2019.

\bibitem{jang2022sequential}
Sooyong Jang, Sangdon Park, Insup Lee, and Osbert Bastani.
\newblock Sequential covariate shift detection using classifier two-sample tests.
\newblock In {\em International Conference on Machine Learning}, pages 9845--9880. PMLR, 2022.

\bibitem{johnson2019billion}
Jeff Johnson, Matthijs Douze, and Herv{\'e} J{\'e}gou.
\newblock Billion-scale similarity search with gpus.
\newblock {\em IEEE Transactions on Big Data}, 7(3):535--547, 2019.

\bibitem{katz-etal:CAV:2017:reluplex}
Guy Katz, Clark Barrett, David~L Dill, Kyle Julian, and Mykel~J Kochenderfer.
\newblock Reluplex: An efficient smt solver for verifying deep neural networks.
\newblock In {\em Computer Aided Verification: 29th International Conference, CAV 2017, Heidelberg, Germany, July 24-28, 2017, Proceedings, Part I 30}, pages 97--117. Springer, 2017.

\bibitem{DBLP:conf/cav/KatzHIJLLSTWZDK19}
Guy Katz, Derek~A. Huang, Duligur Ibeling, Kyle Julian, Christopher Lazarus, Rachel Lim, Parth Shah, Shantanu Thakoor, Haoze Wu, Aleksandar Zeljic, David~L. Dill, Mykel~J. Kochenderfer, and Clark~W. Barrett.
\newblock The marabou framework for verification and analysis of deep neural networks.
\newblock In Isil Dillig and Serdar Tasiran, editors, {\em Computer Aided Verification - 31st International Conference, {CAV} 2019, New York City, NY, USA, July 15-18, 2019, Proceedings, Part {I}}, volume 11561 of {\em Lecture Notes in Computer Science}, pages 443--452. Springer, 2019.

\bibitem{kay2015unequal}
Matthew Kay, Cynthia Matuszek, and Sean~A Munson.
\newblock Unequal representation and gender stereotypes in image search results for occupations.
\newblock In {\em Proceedings of the 33rd annual acm conference on human factors in computing systems}, pages 3819--3828, 2015.

\bibitem{kazmierczak2024MLOps}
Jarek Kazmierczak, Khalid Salama, and Valentin Huerta.
\newblock Mlops: Continuous delivery and automation pipelines in machine learning.
\newblock \url{https://cloud.google.com/architecture/mlops-continuous-delivery-and-automation-pipelines-in-machine-learning}, 2024.
\newblock Accessed: 2024-12-16.

\bibitem{kendall2017uncertainties}
Alex Kendall and Yarin Gal.
\newblock What uncertainties do we need in bayesian deep learning for computer vision?
\newblock {\em Advances in neural information processing systems}, 30, 2017.

\bibitem{kirchenbauer2022your}
John Kirchenbauer, Jacob Oaks, and Eric Heim.
\newblock What is your metric telling you? evaluating classifier calibration under context-specific definitions of reliability.
\newblock {\em arXiv preprint arXiv:2205.11454}, 2022.

\bibitem{koh2021wilds}
Pang~Wei Koh, Shiori Sagawa, Henrik Marklund, Sang~Michael Xie, Marvin Zhang, Akshay Balsubramani, Weihua Hu, Michihiro Yasunaga, Richard~Lanas Phillips, Irena Gao, et~al.
\newblock Wilds: A benchmark of in-the-wild distribution shifts.
\newblock In {\em International conference on machine learning}, pages 5637--5664. PMLR, 2021.

\bibitem{koopman2019many}
Philip Koopman, Frank Fratrik, et~al.
\newblock How many operational design domains, objects, and events?
\newblock {\em Safeai@ aaai}, 4(4), 2019.

\bibitem{kotha2023provably}
Suhas Kotha, Christopher Brix, J.~Zico Kolter, Krishnamurthy Dvijotham, and Huan Zhang.
\newblock Provably bounding neural network preimages.
\newblock In A.~Oh, T.~Neumann, A.~Globerson, K.~Saenko, M.~Hardt, and S.~Levine, editors, {\em Advances in Neural Information Processing Systems}, volume~36, pages 80270--80290. Curran Associates, Inc., 2023.

\bibitem{kreuzberger2023machine}
Dominik Kreuzberger, Niklas K{\"u}hl, and Sebastian Hirschl.
\newblock Machine learning operations (mlops): Overview, definition, and architecture.
\newblock {\em IEEE access}, 11:31866--31879, 2023.

\bibitem{krizhevsky2012imagenet}
Alex Krizhevsky, Ilya Sutskever, and Geoffrey~E Hinton.
\newblock Imagenet classification with deep convolutional neural networks.
\newblock {\em Advances in neural information processing systems}, 25, 2012.

\bibitem{kuhn2010practical}
D~Richard Kuhn, Raghu~N Kacker, Yu~Lei, et~al.
\newblock Practical combinatorial testing.
\newblock {\em NIST special Publication}, 800(142):142, 2010.

\bibitem{kuhn1955hungarian}
Harold~W Kuhn.
\newblock The hungarian method for the assignment problem.
\newblock {\em Naval research logistics quarterly}, 2(1-2):83--97, 1955.

\bibitem{kuleshov2018accurate}
Volodymyr Kuleshov, Nathan Fenner, and Stefano Ermon.
\newblock Accurate uncertainties for deep learning using calibrated regression.
\newblock In {\em International conference on machine learning}, pages 2796--2804. PMLR, 2018.

\bibitem{kumar2019verified}
Ananya Kumar, Percy~S Liang, and Tengyu Ma.
\newblock Verified uncertainty calibration.
\newblock {\em Advances in Neural Information Processing Systems}, 32, 2019.

\bibitem{kumar2022fine}
Ananya Kumar, Aditi Raghunathan, Robbie Jones, Tengyu Ma, and Percy Liang.
\newblock Fine-tuning can distort pretrained features and underperform out-of-distribution.
\newblock In {\em International Conference on Learning Representations}, 2022.

\bibitem{lakshminarayanan2017simple}
Balaji Lakshminarayanan, Alexander Pritzel, and Charles Blundell.
\newblock Simple and scalable predictive uncertainty estimation using deep ensembles.
\newblock {\em Advances in neural information processing systems}, 30, 2017.

\bibitem{lee2018simple}
Kimin Lee, Kibok Lee, Honglak Lee, and Jinwoo Shin.
\newblock A simple unified framework for detecting out-of-distribution samples and adversarial attacks.
\newblock {\em Advances in neural information processing systems}, 31, 2018.

\bibitem{lee2023surgical}
Yoonho Lee, Annie~S Chen, Fahim Tajwar, Ananya Kumar, Huaxiu Yao, Percy Liang, and Chelsea Finn.
\newblock Surgical fine-tuning improves adaptation to distribution shifts.
\newblock In {\em The Eleventh International Conference on Learning Representations}, 2023.

\bibitem{lei2014distribution}
Jing Lei and Larry Wasserman.
\newblock Distribution-free prediction bands for non-parametric regression.
\newblock {\em Journal of the Royal Statistical Society Series B: Statistical Methodology}, 76(1):71--96, 2014.

\bibitem{leng2024taming}
Jixuan Leng, Chengsong Huang, Banghua Zhu, and Jiaxin Huang.
\newblock Taming overconfidence in llms: Reward calibration in rlhf.
\newblock {\em arXiv preprint arXiv:2410.09724}, 2024.

\bibitem{liang2018enhancing}
Shiyu Liang, Yixuan Li, and R~Srikant.
\newblock Enhancing the reliability of out-of-distribution image detection in neural networks.
\newblock In {\em International Conference on Learning Representations}, 2018.

\bibitem{lipton2018detecting}
Zachary Lipton, Yu-Xiang Wang, and Alexander Smola.
\newblock Detecting and correcting for label shift with black box predictors.
\newblock In {\em International conference on machine learning}, pages 3122--3130. PMLR, 2018.

\bibitem{lipton2023reliable}
Zachary~Chase Lipton.
\newblock Reliable deep learning in dynamic environments.
\newblock In {\em Medical Imaging 2023: Computer-Aided Diagnosis}, volume 12465, page 1246502. SPIE, 2023.

\bibitem{liu2020energy}
Weitang Liu, Xiaoyun Wang, John Owens, and Yixuan Li.
\newblock Energy-based out-of-distribution detection.
\newblock {\em Advances in neural information processing systems}, 33:21464--21475, 2020.

\bibitem{DBLP:conf/cav/LopezCTJ23}
Diego~Manzanas Lopez, Sung~Woo Choi, Hoang{-}Dung Tran, and Taylor~T. Johnson.
\newblock {NNV} 2.0: The neural network verification tool.
\newblock In Constantin Enea and Akash Lal, editors, {\em Computer Aided Verification - 35th International Conference, {CAV} 2023, Paris, France, July 17-22, 2023, Proceedings, Part {II}}, volume 13965 of {\em Lecture Notes in Computer Science}, pages 397--412. Springer, 2023.

\bibitem{maddox2019simple}
Wesley~J Maddox, Pavel Izmailov, Timur Garipov, Dmitry~P Vetrov, and Andrew~Gordon Wilson.
\newblock A simple baseline for bayesian uncertainty in deep learning.
\newblock {\em Advances in neural information processing systems}, 32, 2019.

\bibitem{magee2016spatial}
Lucas Magee, Lee~M Seversky, and Eric Heim.
\newblock Spatial active learning for cost-effective sensing and feature extraction.
\newblock {\em ICML Workshop on Data-Efficient Machine Learning}, 2016.

\bibitem{mani2019coverage}
Senthil Mani, Anush Sankaran, Srikanth Tamilselvam, and Akshay Sethi.
\newblock Coverage testing of deep learning models using dataset characterization.
\newblock {\em arXiv preprint arXiv:1911.07309}, 2019.

\bibitem{mao2024two}
Anqi Mao, Christopher Mohri, Mehryar Mohri, and Yutao Zhong.
\newblock Two-stage learning to defer with multiple experts.
\newblock {\em Advances in neural information processing systems}, 36, 2024.

\bibitem{mccarthy_1969_philosophical}
John McCarthy and Patrick Hayes.
\newblock Some philosophical problems from the standpoint of artificial intelligence.
\newblock In B.~Meltzer and Donald Michie, editors, {\em Machine Intelligence 4}, pages 463--502. Edinburgh University Press, 1969.

\bibitem{mckeeman1998differential}
William~M McKeeman.
\newblock Differential testing for software.
\newblock {\em Digital Technical Journal}, 10(1):100--107, 1998.

\bibitem{miller2021accuracy}
John~P Miller, Rohan Taori, Aditi Raghunathan, Shiori Sagawa, Pang~Wei Koh, Vaishaal Shankar, Percy Liang, Yair Carmon, and Ludwig Schmidt.
\newblock Accuracy on the line: on the strong correlation between out-of-distribution and in-distribution generalization.
\newblock In {\em International conference on machine learning}, pages 7721--7735. PMLR, 2021.

\bibitem{minderer2021revisiting}
Matthias Minderer, Josip Djolonga, Rob Romijnders, Frances Hubis, Xiaohua Zhai, Neil Houlsby, Dustin Tran, and Mario Lucic.
\newblock Revisiting the calibration of modern neural networks.
\newblock {\em Advances in Neural Information Processing Systems}, 34:15682--15694, 2021.

\bibitem{ming2022impact}
Yifei Ming, Hang Yin, and Yixuan Li.
\newblock On the impact of spurious correlation for out-of-distribution detection.
\newblock In {\em Proceedings of the AAAI conference on artificial intelligence}, volume~36, pages 10051--10059, 2022.

\bibitem{morris2020textattack}
John Morris, Eli Lifland, Jin~Yong Yoo, Jake Grigsby, Di~Jin, and Yanjun Qi.
\newblock Textattack: A framework for adversarial attacks, data augmentation, and adversarial training in nlp.
\newblock In {\em Proceedings of the 2020 Conference on Empirical Methods in Natural Language Processing: System Demonstrations}, pages 119--126, 2020.

\bibitem{mosser2022comprehensive}
Lukas Mosser and Ehsan Zabihi~Naeini.
\newblock A comprehensive study of calibration and uncertainty quantification for bayesian convolutional neural networks—an application to seismic data.
\newblock {\em Geophysics}, 87(4):IM157--IM176, 2022.

\bibitem{mukhoti2020calibrating}
Jishnu Mukhoti, Viveka Kulharia, Amartya Sanyal, Stuart Golodetz, Philip Torr, and Puneet Dokania.
\newblock Calibrating deep neural networks using focal loss.
\newblock {\em Advances in Neural Information Processing Systems}, 33:15288--15299, 2020.

\bibitem{murphy2022probabilistic}
Kevin~P Murphy.
\newblock {\em Probabilistic machine learning: an introduction}.
\newblock MIT press, 2022.

\bibitem{naeini2015obtaining}
Mahdi~Pakdaman Naeini, Gregory Cooper, and Milos Hauskrecht.
\newblock Obtaining well calibrated probabilities using bayesian binning.
\newblock In {\em Proceedings of the AAAI conference on artificial intelligence}, volume~29, 2015.

\bibitem{neal1992belief}
Radford~M. Neal.
\newblock Connectionist learning of belief networks.
\newblock {\em Artificial Intelligence}, 56(1):71--113, 1992.

\bibitem{ng2018machine}
Andrew Ng.
\newblock Machine learning yearning: Technical strategy for ai engineers, in the era of deep learning.
\newblock 2018.

\bibitem{nguyen2015deep}
Anh Nguyen, Jason Yosinski, and Jeff Clune.
\newblock Deep neural networks are easily fooled: High confidence predictions for unrecognizable images.
\newblock In {\em Proceedings of the IEEE conference on computer vision and pattern recognition}, pages 427--436, 2015.

\bibitem{nixon2019measuring}
Jeremy Nixon, Michael~W Dusenberry, Linchuan Zhang, Ghassen Jerfel, and Dustin Tran.
\newblock Measuring calibration in deep learning.
\newblock In {\em CVPR workshops}, volume~2, 2019.

\bibitem{o2012practical}
Patrick O'Connor and Andre Kleyner.
\newblock {\em Practical reliability engineering}.
\newblock John Wiley \& Sons, 2012.

\bibitem{OpenAIData}
OpenAI.
\newblock Our approach to data and ai.
\newblock \url{https://openai.com/index/approach-to-data-and-ai/}, 2024.
\newblock Accessed: 2024-11-07.

\bibitem{oquab2024dinov2}
Maxime Oquab, Timoth{\'e}e Darcet, Th{\'e}o Moutakanni, Huy Vo, Marc Szafraniec, Vasil Khalidov, Pierre Fernandez, Daniel Haziza, Francisco Massa, Alaaeldin El-Nouby, et~al.
\newblock Dinov2: Learning robust visual features without supervision.
\newblock {\em Transactions on Machine Learning Research Journal}, pages 1--31, 2024.

\bibitem{padilla2020survey}
Rafael Padilla, Sergio~L Netto, and Eduardo~AB Da~Silva.
\newblock A survey on performance metrics for object-detection algorithms.
\newblock In {\em 2020 international conference on systems, signals and image processing (IWSSIP)}, pages 237--242. IEEE, 2020.

\bibitem{papernot2016limitations}
Nicolas Papernot, Patrick McDaniel, Somesh Jha, Matt Fredrikson, Z~Berkay Celik, and Ananthram Swami.
\newblock The limitations of deep learning in adversarial settings.
\newblock In {\em 2016 IEEE European symposium on security and privacy (EuroS\&P)}, pages 372--387. IEEE, 2016.

\bibitem{parthasarathy2024ultimate}
Venkatesh~Balavadhani Parthasarathy, Ahtsham Zafar, Aafaq Khan, and Arsalan Shahid.
\newblock The ultimate guide to fine-tuning llms from basics to breakthroughs: An exhaustive review of technologies, research, best practices, applied research challenges and opportunities.
\newblock {\em arXiv preprint arXiv:2408.13296}, 2024.

\bibitem{pei2022requirements}
Zhongyi Pei, Lin Liu, Chen Wang, and Jianmin Wang.
\newblock Requirements engineering for machine learning: A review and reflection.
\newblock In {\em 2022 IEEE 30th International Requirements Engineering Conference Workshops (REW)}, pages 166--175. IEEE, 2022.

\bibitem{pesaranghader2016fast}
Ali Pesaranghader and Herna~L Viktor.
\newblock Fast hoeffding drift detection method for evolving data streams.
\newblock In {\em Machine Learning and Knowledge Discovery in Databases: European Conference, ECML PKDD 2016, Riva del Garda, Italy, September 19-23, 2016, Proceedings, Part II 16}, pages 96--111. Springer, 2016.

\bibitem{pezzementi2018putting}
Zachary Pezzementi, Trenton Tabor, Samuel Yim, Jonathan~K Chang, Bill Drozd, David Guttendorf, Michael Wagner, and Philip Koopman.
\newblock Putting image manipulations in context: robustness testing for safe perception.
\newblock In {\em 2018 IEEE International Symposium on Safety, Security, and Rescue Robotics (SSRR)}, pages 1--8. IEEE, 2018.

\bibitem{pietrantuono2010online}
Roberto Pietrantuono, Stefano Russo, and Kishor~S Trivedi.
\newblock Online monitoring of software system reliability.
\newblock In {\em 2010 European Dependable Computing Conference}, pages 209--218. IEEE, 2010.

\bibitem{platt1999probabilistic}
John Platt.
\newblock Probabilistic outputs for support vector machines and comparisons to regularized likelihood methods.
\newblock {\em Advances in large margin classifiers}, 10(3):61--74, 1999.

\bibitem{prabhudesai2023understanding}
Snehal Prabhudesai, Leyao Yang, Sumit Asthana, Xun Huan, Q~Vera Liao, and Nikola Banovic.
\newblock Understanding uncertainty: how lay decision-makers perceive and interpret uncertainty in human-ai decision making.
\newblock In {\em Proceedings of the 28th international conference on intelligent user interfaces}, pages 379--396, 2023.

\bibitem{pykes2023aguide}
Kurtis Pykes.
\newblock A guide to monitoring machine learning models in production.
\newblock \url{https://developer.nvidia.com/blog/a-guide-to-monitoring-machine-learning-models-in-production/}, 2023.
\newblock Accessed: 2024-12-08.

\bibitem{ieee_1990}
Jane Radatz.
\newblock {IEEE} {Standard} {Glossary} of {Software} {Engineering} {Terminology}.
\newblock {\em IEEE Std 610.12-1990}, pages 1--84, December 1990.
\newblock Conference Name: IEEE Std 610.12-1990.

\bibitem{rao2023studying}
Adrit Rao, Joon-Young Lee, and Oliver Aalami.
\newblock Studying the impact of augmentations on medical confidence calibration.
\newblock In {\em Proceedings of the IEEE/CVF International Conference on Computer Vision}, pages 2462--2472, 2023.

\bibitem{ravi2024sam}
Nikhila Ravi, Valentin Gabeur, Yuan-Ting Hu, Ronghang Hu, Chaitanya Ryali, Tengyu Ma, Haitham Khedr, Roman R{\"a}dle, Chloe Rolland, Laura Gustafson, et~al.
\newblock Sam 2: Segment anything in images and videos.
\newblock {\em arXiv preprint arXiv:2408.00714}, 2024.

\bibitem{redmon2016you}
Joseph Redmon, Santosh Divvala, Ross Girshick, and Ali Farhadi.
\newblock You only look once: Unified, real-time object detection.
\newblock In {\em Proceedings of the IEEE conference on computer vision and pattern recognition}, pages 779--788, 2016.

\bibitem{ren2021survey}
Pengzhen Ren, Yun Xiao, Xiaojun Chang, Po-Yao Huang, Zhihui Li, Brij~B Gupta, Xiaojiang Chen, and Xin Wang.
\newblock A survey of deep active learning.
\newblock {\em ACM computing surveys (CSUR)}, 54(9):1--40, 2021.

\bibitem{rezatofighi2019generalized}
Hamid Rezatofighi, Nathan Tsoi, JunYoung Gwak, Amir Sadeghian, Ian Reid, and Silvio Savarese.
\newblock Generalized intersection over union: A metric and a loss for bounding box regression.
\newblock In {\em Proceedings of the IEEE/CVF conference on computer vision and pattern recognition}, pages 658--666, 2019.

\bibitem{ross2021nist}
Ron Ross, Victoria Pillitteri, Richard Graubart, Deborah~J Bodeau, and Rosalie~M McQuaid.
\newblock Nist special publication 800-160, volume 2 revision 1: Developing cyber-resilient systems: A systems security engineering approach.
\newblock In {\em National Institute of Standards and Technology (US)}, number NIST SP 800-160, Vol. 2, Rev. 1; National Institute of Standards and Technology Special Publication 800-160, Vol. 2, Rev. 1. National Institute of Standards and Technology (US), 2021.

\bibitem{roy2001toward}
Nicholas Roy and Andrew McCallum.
\newblock Toward optimal active learning through monte carlo estimation of error reduction.
\newblock 2001.

\bibitem{rudin2019stop}
Cynthia Rudin.
\newblock Stop explaining black box machine learning models for high stakes decisions and use interpretable models instead.
\newblock {\em Nature machine intelligence}, 1(5):206--215, 2019.

\bibitem{russakovsky2015scaling}
Olga Russakovsky.
\newblock {\em Scaling Up Object Detection}.
\newblock Stanford University, 2015.

\bibitem{saerens2002adjusting}
Marco Saerens, Patrice Latinne, and Christine Decaestecker.
\newblock Adjusting the outputs of a classifier to new a priori probabilities: a simple procedure.
\newblock {\em Neural computation}, 14(1):21--41, 2002.

\bibitem{salman2019convex}
Hadi Salman, Greg Yang, Huan Zhang, Cho-Jui Hsieh, and Pengchuan Zhang.
\newblock A convex relaxation barrier to tight robustness verification of neural networks.
\newblock {\em Advances in Neural Information Processing Systems}, 32:9835--9846, 2019.

\bibitem{sener2018active}
Ozan Sener and Silvio Savarese.
\newblock Active learning for convolutional neural networks: A core-set approach.
\newblock In {\em International Conference on Learning Representations}, 2018.

\bibitem{settles2009active}
Burr Settles.
\newblock Active learning literature survey.
\newblock 2009.

\bibitem{shafer2008tutorial}
Glenn Shafer and Vladimir Vovk.
\newblock A tutorial on conformal prediction.
\newblock {\em Journal of Machine Learning Research}, 9(3), 2008.

\bibitem{sharif2016accessorize}
Mahmood Sharif, Sruti Bhagavatula, Lujo Bauer, and Michael~K Reiter.
\newblock Accessorize to a crime: Real and stealthy attacks on state-of-the-art face recognition.
\newblock In {\em Proceedings of the 2016 acm sigsac conference on computer and communications security}, pages 1528--1540, 2016.

\bibitem{shi2024genbab}
Zhouxing Shi, Qirui Jin, Zico Kolter, Suman Jana, Cho-Jui Hsieh, and Huan Zhang.
\newblock Neural network verification with branch-and-bound for general nonlinearities.
\newblock {\em arXiv preprint arXiv:2405.21063}, 2024.

\bibitem{shimodaira2000improving}
Hidetoshi Shimodaira.
\newblock Improving predictive inference under covariate shift by weighting the log-likelihood function.
\newblock {\em Journal of statistical planning and inference}, 90(2):227--244, 2000.

\bibitem{simonite2018when}
Tom Simonite.
\newblock When it comes to gorillas, google photos remains blind.
\newblock \url{https://www.wired.com/story/when-it-comes-to-gorillas-google-photos-remains-blind/}, 2018.
\newblock Accessed: 2024-11-07.

\bibitem{DBLP:journals/pacmpl/SinghGPV19}
Gagandeep Singh, Timon Gehr, Markus P{\"{u}}schel, and Martin~T. Vechev.
\newblock An abstract domain for certifying neural networks.
\newblock {\em Proc. {ACM} Program. Lang.}, 3({POPL}):41:1--41:30, 2019.

\bibitem{sinha2018certifying}
Aman Sinha, Hongseok Namkoong, and John Duchi.
\newblock Certifying some distributional robustness with principled adversarial training.
\newblock In {\em International Conference on Learning Representations}, 2018.

\bibitem{stojanov2019low}
Petar Stojanov, Mingming Gong, Jaime Carbonell, and Kun Zhang.
\newblock Low-dimensional density ratio estimation for covariate shift correction.
\newblock In {\em The 22nd international conference on artificial intelligence and statistics}, pages 3449--3458. PMLR, 2019.

\bibitem{storkey2008training}
Amos Storkey, J~Qui{\~n}onero-Candela, M~Sugiyama, A~Schwaighofer, and ND~Lawrence.
\newblock When training and test sets are different: Characterizing learning transfer.
\newblock In {\em Dataset Shift in Machine Learning}, pages 3--28. Yale University Press in association with the Museum of London, 2008.

\bibitem{suarez2023survey}
Andr{\'e}s~L Su{\'a}rez-Cetrulo, David Quintana, and Alejandro Cervantes.
\newblock A survey on machine learning for recurring concept drifting data streams.
\newblock {\em Expert Systems with Applications}, 213:118934, 2023.

\bibitem{sugiyama2007direct}
Masashi Sugiyama, Shinichi Nakajima, Hisashi Kashima, Paul Buenau, and Motoaki Kawanabe.
\newblock Direct importance estimation with model selection and its application to covariate shift adaptation.
\newblock {\em Advances in neural information processing systems}, 20, 2007.

\bibitem{sugiyama2012density}
Masashi Sugiyama, Taiji Suzuki, and Takafumi Kanamori.
\newblock Density-ratio matching under the bregman divergence: a unified framework of density-ratio estimation.
\newblock {\em Annals of the Institute of Statistical Mathematics}, 64:1009--1044, 2012.

\bibitem{suh2024survey}
Namjoon Suh and Guang Cheng.
\newblock A survey on statistical theory of deep learning: Approximation, training dynamics, and generative models.
\newblock {\em arXiv preprint arXiv:2401.07187}, 2024.

\bibitem{sun2024evaluating}
Jiuding Sun, Chantal Shaib, and Byron~C Wallace.
\newblock Evaluating the zero-shot robustness of instruction-tuned language models.
\newblock In {\em The Twelfth International Conference on Learning Representations}, 2024.

\bibitem{sun2021react}
Yiyou Sun, Chuan Guo, and Yixuan Li.
\newblock React: Out-of-distribution detection with rectified activations.
\newblock {\em Advances in Neural Information Processing Systems}, 34:144--157, 2021.

\bibitem{sun2022out}
Yiyou Sun, Yifei Ming, Xiaojin Zhu, and Yixuan Li.
\newblock Out-of-distribution detection with deep nearest neighbors.
\newblock In {\em International Conference on Machine Learning}, pages 20827--20840. PMLR, 2022.

\bibitem{sun2019structural}
Youcheng Sun, Xiaowei Huang, Daniel Kroening, James Sharp, Matthew Hill, and Rob Ashmore.
\newblock Structural test coverage criteria for deep neural networks.
\newblock {\em ACM Transactions on Embedded Computing Systems (TECS)}, 18(5s):1--23, 2019.

\bibitem{szegedy2014}
Christian Szegedy, Wojciech Zaremba, Ilya Sutskever, Joan Bruna, Dumitru Erhan, Ian Goodfellow, and Rob Fergus.
\newblock Intriguing properties of neural networks.
\newblock In {\em International Conference on Learning Representations}, 2014.

\bibitem{tao2023dual}
Linwei Tao, Minjing Dong, and Chang Xu.
\newblock Dual focal loss for calibration.
\newblock In {\em International Conference on Machine Learning}, pages 33833--33849. PMLR, 2023.

\bibitem{touvron2023llama}
Hugo Touvron, Louis Martin, Kevin Stone, Peter Albert, Amjad Almahairi, Yasmine Babaei, Nikolay Bashlykov, Soumya Batra, Prajjwal Bhargava, Shruti Bhosale, et~al.
\newblock Llama 2: Open foundation and fine-tuned chat models.
\newblock {\em arXiv preprint arXiv:2307.09288}, 2023.

\bibitem{touvron2023llama2}
Hugo Touvron, Louis Martin, Kevin Stone, Peter Albert, Amjad Almahairi, Yasmine Babaei, Nikolay Bashlykov, Soumya Batra, Prajjwal Bhargava, Shruti Bhosale, et~al.
\newblock Llama 2: Open foundation and fine-tuned chat models.
\newblock {\em arXiv preprint arXiv:2307.09288}, 2023.

\bibitem{tu2020physically}
James Tu, Mengye Ren, Sivabalan Manivasagam, Ming Liang, Bin Yang, Richard Du, Frank Cheng, and Raquel Urtasun.
\newblock Physically realizable adversarial examples for lidar object detection.
\newblock In {\em Proceedings of the IEEE/CVF conference on computer vision and pattern recognition}, pages 13716--13725, 2020.

\bibitem{vapnik2013nature}
Vladimir Vapnik.
\newblock {\em The nature of statistical learning theory}.
\newblock Springer science \& business media, 2013.

\bibitem{villamizar2021requirements}
Hugo Villamizar, Tatiana Escovedo, and Marcos Kalinowski.
\newblock Requirements engineering for machine learning: A systematic mapping study.
\newblock In {\em 2021 47th Euromicro Conference on Software Engineering and Advanced Applications (SEAA)}, pages 29--36. IEEE, 2021.

\bibitem{visengeriyeva2024MLOps}
Larysa Visengeriyeva, Isabel Kammer, Anja amd~Bär, Alexander Kniesz, and Michael Plöd.
\newblock Mlops principles.
\newblock \url{https://ml-ops.org/content/mlops-principles}.
\newblock Accessed: 2024-12-16.

\bibitem{vogelsang2019requirements}
Andreas Vogelsang and Markus Borg.
\newblock Requirements engineering for machine learning: Perspectives from data scientists.
\newblock In {\em 2019 IEEE 27th International Requirements Engineering Conference Workshops (REW)}, pages 245--251. IEEE, 2019.

\bibitem{wang2021beta}
Shiqi Wang, Huan Zhang, Kaidi Xu, Xue Lin, Suman Jana, Cho-Jui Hsieh, and J~Zico Kolter.
\newblock {Beta-CROWN}: Efficient bound propagation with per-neuron split constraints for complete and incomplete neural network verification.
\newblock {\em Advances in Neural Information Processing Systems}, 34, 2021.

\bibitem{wen2021combining}
Yeming Wen, Ghassen Jerfel, Rafael Muller, Michael~W Dusenberry, Jasper Snoek, Balaji Lakshminarayanan, and Dustin Tran.
\newblock Combining ensembles and data augmentation can harm your calibration.
\newblock In {\em International Conference on Learning Representations}, 2021.

\bibitem{weyns2023towards}
Danny Weyns, Radu Calinescu, Raffaela Mirandola, Kenji Tei, Maribel Acosta, Nelly Bencomo, Amel Bennaceur, Nicolas Boltz, Tomas Bures, Javier Camara, et~al.
\newblock Towards a research agenda for understanding and managing uncertainty in self-adaptive systems.
\newblock {\em ACM SIGSOFT Software Engineering Notes}, 48(4):20--36, 2023.

\bibitem{wright2024analytic}
Oren Wright, Yorie Nakahira, and Jos{\'e}~MF Moura.
\newblock An analytic solution to covariance propagation in neural networks.
\newblock In {\em International Conference on Artificial Intelligence and Statistics}, pages 4087--4095. PMLR, 2024.

\bibitem{wymore2018model}
A~Wayne Wymore.
\newblock {\em Model-based systems engineering}.
\newblock CRC press, 2018.

\bibitem{xiang2021patchguard}
Chong Xiang, Arjun~Nitin Bhagoji, Vikash Sehwag, and Prateek Mittal.
\newblock $\{$PatchGuard$\}$: A provably robust defense against adversarial patches via small receptive fields and masking.
\newblock In {\em 30th USENIX Security Symposium (USENIX Security 21)}, pages 2237--2254, 2021.

\bibitem{xie2024importance}
Renchunzi Xie, Hongxin Wei, Lei Feng, Yuzhou Cao, and Bo~An.
\newblock On the importance of feature separability in predicting out-of-distribution error.
\newblock {\em Advances in Neural Information Processing Systems}, 36, 2024.

\bibitem{xu2020automatic}
Kaidi Xu, Zhouxing Shi, Huan Zhang, Yihan Wang, Kai-Wei Chang, Minlie Huang, Bhavya Kailkhura, Xue Lin, and Cho-Jui Hsieh.
\newblock Automatic perturbation analysis for scalable certified robustness and beyond.
\newblock {\em Advances in Neural Information Processing Systems}, 33, 2020.

\bibitem{xu2021fast}
Kaidi Xu, Huan Zhang, Shiqi Wang, Yihan Wang, Suman Jana, Xue Lin, and Cho-Jui Hsieh.
\newblock {Fast and Complete}: Enabling complete neural network verification with rapid and massively parallel incomplete verifiers.
\newblock In {\em International Conference on Learning Representations}, 2021.

\bibitem{aiid:46}
Roman Yampolskiy.
\newblock Incident number 46.
\newblock {\em AI Incident Database}, 2014.

\bibitem{yuan2010classification}
Ming Yuan and Marten Wegkamp.
\newblock Classification methods with reject option based on convex risk minimization.
\newblock {\em Journal of Machine Learning Research}, 11(1), 2010.

\bibitem{zadrozny2004learning}
Bianca Zadrozny.
\newblock Learning and evaluating classifiers under sample selection bias.
\newblock In {\em Proceedings of the twenty-first international conference on Machine learning}, page 114, 2004.

\bibitem{zadrozny2001obtaining}
Bianca Zadrozny and Charles Elkan.
\newblock Obtaining calibrated probability estimates from decision trees and naive bayesian classifiers.
\newblock In {\em International Conference on Machine Learning}, volume~1, pages 609--616, 2001.

\bibitem{zadrozny2002transforming}
Bianca Zadrozny and Charles Elkan.
\newblock Transforming classifier scores into accurate multiclass probability estimates.
\newblock In {\em Proceedings of the eighth ACM SIGKDD international conference on Knowledge discovery and data mining}, pages 694--699, 2002.

\bibitem{zhang_theoretically_2019}
Hongyang Zhang, Yaodong Yu, Jiantao Jiao, Eric Xing, Laurent~El Ghaoui, and Michael Jordan.
\newblock Theoretically {Principled} {Trade}-off between {Robustness} and {Accuracy}.
\newblock In {\em Proceedings of the 36th {International} {Conference} on {Machine} {Learning}}, pages 7472--7482. PMLR, May 2019.
\newblock ISSN: 2640-3498.

\bibitem{zhang2022general}
Huan Zhang, Shiqi Wang, Kaidi Xu, Linyi Li, Bo~Li, Suman Jana, Cho-Jui Hsieh, and J~Zico Kolter.
\newblock General cutting planes for bound-propagation-based neural network verification.
\newblock {\em Advances in Neural Information Processing Systems}, 2022.

\bibitem{zhang22babattack}
Huan Zhang, Shiqi Wang, Kaidi Xu, Yihan Wang, Suman Jana, Cho-Jui Hsieh, and Zico Kolter.
\newblock A branch and bound framework for stronger adversarial attacks of {R}e{LU} networks.
\newblock In {\em Proceedings of the 39th International Conference on Machine Learning}, volume 162, pages 26591--26604, 2022.

\bibitem{zhang2018efficient}
Huan Zhang, Tsui-Wei Weng, Pin-Yu Chen, Cho-Jui Hsieh, and Luca Daniel.
\newblock Efficient neural network robustness certification with general activation functions.
\newblock {\em Advances in Neural Information Processing Systems}, 31:4939--4948, 2018.

\bibitem{zhang2024adapting}
Yu-Jie Zhang, Zhen-Yu Zhang, Peng Zhao, and Masashi Sugiyama.
\newblock Adapting to continuous covariate shift via online density ratio estimation.
\newblock {\em Advances in Neural Information Processing Systems}, 36, 2024.

\bibitem{zhao2020maximum}
Long Zhao, Ting Liu, Xi~Peng, and Dimitris Metaxas.
\newblock Maximum-entropy adversarial data augmentation for improved generalization and robustness.
\newblock {\em Advances in Neural Information Processing Systems}, 33:14435--14447, 2020.

\bibitem{zou2023universal}
Andy Zou, Zifan Wang, J.~Zico Kolter, and Matt Fredrikson.
\newblock Universal and transferable adversarial attacks on aligned language models, 2023.

\end{thebibliography}
\end{document}